\documentclass[10pt,twocolumn,letterpaper]{article}

\usepackage[pagenumbers]{cvpr} % To force page numbers, e.g. for an arXiv version
\usepackage{cuted}    % for strip environment
\usepackage{capt-of}  % for \captionof outside floats
\usepackage{booktabs}
\usepackage{multirow}

\newcommand{\OURS}{PercHead}
\definecolor{cvprblue}{rgb}{0.21,0.49,0.74}
\usepackage[pagebackref,breaklinks,colorlinks,allcolors=cvprblue]{hyperref}

\def\paperID{} % *** Enter the Paper ID here
\def\confName{CVPR}
\def\confYear{2026}

\title{{\OURS}: Perceptual Head Model \\ for Single-Image 3D Head Reconstruction \& Editing}

\author{
Antonio Oroz \qquad Matthias Nie{\ss}ner \qquad Tobias Kirschstein\\
Technical University of Munich\\
{\tt\small \{antonio.oroz, niessner, tobias.kirschstein\}@tum.de}
}

\begin{document}
\maketitle

\begin{strip}
\vspace{-1cm}
    \centering
    \includegraphics[width=\textwidth]{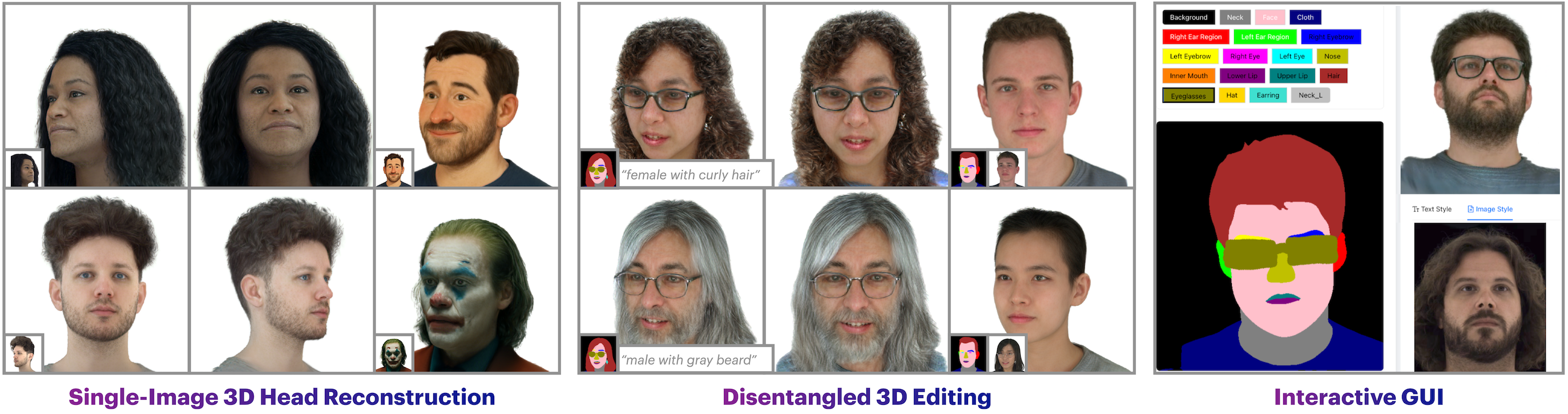}
    \captionof{figure}{\textbf{{\OURS}.} Our method reconstructs high-fidelity 3D heads from single input images, maintaining consistency across arbitrary viewpoints. In our interactive GUI, a second fine-tuned editing model generates realistic 3D heads from a segmentation-map, with style controlled through either a reference image or a text prompt.
}
    \label{fig:teaser}
\end{strip}

\begin{abstract}
We present {\OURS}, a model for single-image 3D head reconstruction and disentangled 3D editing - two tasks that are inherently challenging due to ambiguity in plausible explanations for the same input. At the heart of our approach lies our novel perceptual loss based on DINOv2 \cite{dinov2} and SAM 2.1 \cite{sam2}. Unlike widely-adopted low-level losses like LPIPS \cite{LPIPS}, SSIM \cite{SSIM} or L1, we rely on deep visual understanding of images and the resulting generalized supervision signals. We show that our new loss can be a drop-in replacement for standard losses and used to improve visual quality in high-frequency areas. We base our model architecture on Vision Transformers (ViTs) \cite{vit}, allowing us to decouple the 3D representation from the 2D input. We train our method on multi-view images \cite{nersemble, cafca} for view-consistency and in-the-wild images \cite{ffhq} for strong transferability to new environments. Our model achieves state-of-the-art performance in novel-view synthesis and, furthermore, exhibits exceptional robustness to extreme viewing angles. We also extend our base model to disentangled 3D editing by swapping the encoder and fine-tuning the network. A segmentation map controls geometry and either a text prompt or a reference image specifies appearance. We highlight the intuitive and powerful 3D editing capabilities through an interactive GUI.

The training and inference code, together with the interactive GUI, is available on the project website: 
\href{https://antoniooroz.github.io/PercHead/}{https://antoniooroz.github.io/PercHead/}
%\vspace{0.0em}

\end{abstract}    
\section{Introduction}
\label{sec:intro}

Creating 3D heads from single inputs is an incredibly interesting area of research, which can enable many exciting applications for virtual conferencing, gaming, or personalized digital avatars. It can also open the door to advanced tasks such as camera control over portraits, 3D appearance editing, relighting under novel illumination conditions, and many more. However, enabling any of the aforementioned scenarios requires models that are not only high-fidelity but also fundamentally 3D-consistent, capable of producing plausible geometry and appearance even in regions never observed in the input view.

Solving these challenges is difficult: from just one image, the 3D reconstruction is inherently ambiguous, permitting many plausible explanations of the same observation. To resolve this, we require strong priors from large multi-view datasets. In practice, such datasets in the head domain are either limited in size \cite{nersemble, ava256} or synthetic \cite{cafca}. More extensive datasets can only be assembled efficiently from in-the-wild data \cite{ffhq, vfhq}, which is inherently single-view or, at best, offers only limited viewpoint diversity. As a consequence, existing methods \cite{gagavatar, lam, panohead, spherehead} tend to perform well for views close to the input and deteriorate significantly when the camera is moved. On top of that, humans are exceptionally sensitive to even small errors in facial structure and appearance, demanding extremely high visual fidelity. To supervise such methods, it's often difficult to rely on low-level image comparisons which will often discard plausible, but not pixel-perfect, reconstructions. For more advanced tasks, such as generative tasks like disentangled editing, this is even more pronounced.

In previous works, these challenges we're aimed to be addressed through various approaches: One of the early methods, ROME \cite{rome}, relies on a strong geometric prior through mesh-based 3D morphable models, offering a great consistency, but with limited realism especially in complex regions such as hair. GAN-based methods~\cite{eg3d, panohead, spherehead} leverage large-scale 2D datasets and adversarial losses, offering great perceptual realism. However, GANs are notoriously difficult to train and often require inversion for reconstruction, which limits identity preservation. GAGAvatar \cite{gagavatar}, on the other hand, uses a dual-lifting Gaussian Splatting pipeline trained on multi-frame video data, leading to highly realistic outputs. The dual-lifting strategy, however, limits 3D consistency, leading to unrealistic outputs when viewing angles differ from the input. The Large Avatar Model (LAM)~\cite{lam} instead adapts a transformer-based architecture, achieving strong reconstructions for frontal inputs but exhibiting similar shortcomings as GAGAvatar. LGM~\cite{lgm} employs a generalized multi-view diffusion prior, which is incredibly versatile, but not able to produce the high-realism outputs required for human head synthesis.

We propose a method that reconstructs highly realistic 3D heads from inputs taken at any viewing angle and renders them with full 3D consistency to even extreme target viewpoints. We also adopt a Vision Transformer (ViT) based architecture, due to its ability to decouple the 3D representation from the 2D input. We carefully construct our training task with a mixture of multi-view datasets \cite{cafca, nersemble}, providing diverse input and output views, and an in-the-wild dataset \cite{ffhq}, which extends the training data to diverse scenarios. Critically, supervising our pipeline with standard low-level losses, such as LPIPS \cite{LPIPS}, SSIM \cite{SSIM}, or L1, would lead to noisy signals in high-frequency areas such as hair. Therefore, we turn to the recent progress in foundation vision models, specifically DINOv2 \cite{dinov2} and SAM 2.1 \cite{sam2}. These new methods offer deep understanding of images, even being able to solve a multitude of tasks without specifically training for them. Consequently, they can also be used to compare rendered and ground truth images on a generalized, perceptual level, leading to more robust training and better visual quality.

Our method significantly surpasses current approaches, such as PanoHead \cite{panohead}, SphereHead \cite{spherehead}, GAGAvatar \cite{gagavatar}, LGM \cite{lgm} and LAM \cite{lam}, particularly in reconstruction tasks involving extreme viewpoints. Furthermore, our generalized 3D reconstruction model serves as a versatile foundation that can be extended to tasks such as 3D editing. By replacing the encoder and fine-tuning the model with segmentation maps~\cite{farl} and CLIP embeddings~\cite{clip} as inputs, we enable advanced generative capabilities that disentangle geometric and style control.

To summarize, our contributions are:

\begin{itemize}

\item A new model which provides state-of-the-art performance for single-image 3D head reconstruction, both for novel and extreme views

\item A loss formulation based solely on deep perceptual supervision, significantly improving quality in highly complex areas like hair

\item Advanced 3D face editing capabilities through an interactive GUI, enabled by a fine-tuned version of our generalized base model

\end{itemize}
\section{Related Work}
\label{sec:related_work}

\subsection{Single-Image 3D Face Reconstruction}

Early single-image 3D reconstruction methods used mesh-based models \cite{faceverse, ffhq_uv, rome, avatarme, avatarmepp, lin2020towards, ganfit, ganfit_fast}, offering efficient geometry but limited detail and flexibility. GAN-based refinements \cite{ganfit, ganfit_fast} improve realism but still inherit the mesh constraints.

Early NeRF-/volume-rendering-based GANs~\cite{pi_gan, style_sdf, style_nerf} achieved high-quality 3D head synthesis but were slow or not completely 3D consistent. EG3D~\cite{eg3d} introduced tri-plane representations to improve efficiency, enabling faster rendering with great visual quality. PanoHead~\cite{panohead} and SphereHead~\cite{spherehead} extended this to 360° head generation. Nonetheless, these GAN-based methods require computationally expensive latent inversion~\cite{pti} for 3D head reconstruction. Encoder-based variants~\cite{triplanenet, mega, dualencoderganinversion} improve inference speed but typically reduce reconstruction fidelity. Most GAN-based methods are also trained on 2D data and lack strong multi-view supervision, limiting 3D consistency. Trevithick et al. \cite{Trevithick} provide a method trained on synthetic data from a pre-trained 3D-aware image generator. However, their results show that their model still struggles with more extreme side-view inputs. Portrait4D~\cite{portrait4d} and Portrait4Dv2~\cite{portrait4d_v2} also address view-consistency by introducing synthetic multi-view supervision, resulting in more consistent geometry. Nonetheless, NeRFs remain computationally heavy and can still suffer from view-dependent variability.

With the emergence of Gaussian Splatting~\cite{gaussian_splatting} as a leading representation for fast, photorealistic 3D rendering, several recent works have adopted it in the 3D head domain~\cite{gghead, gaussian_avatars, hhavatar, npga, avat3r}. However, these methods are either unconditional generative models, subject-specific optimizations, or require multi-view or video input. GAGAvatar~\cite{gagavatar} and LAM~\cite{lam} apply Gaussian Splatting to single-image 3D head reconstructions. While they are highly efficient and produce impressive results for near-input view angles, both deteriorate in performance for more diverse viewpoints.

Building on the popularity of diffusion models~\cite{gu2021vector, stable_diffusion, Hang_2023_ICCV, JMLR:v23:21-0635, podell2023sdxlimprovinglatentdiffusion}, numerous diffusion and SDS-based \cite{sds} methods~\cite{arc2avatar, arc2face, rodin, rodinhd, DiffPortrait3D, gu2025diffportrait360} have emerged. However, they typically operate in image-space, require per-subject optimization, fall short of photorealistic 3D head synthesis, or require large amounts of GPU memory. LGM~\cite{lgm} introduces a multi-view diffusion prior to improve 3D consistency, but is not able to produce the high fidelity required for head synthesis. FaceLift \cite{facelift} trains a multi-view diffusion prior on synthetic heads, but struggles with side views.

Many of the aforementioned methods either use adversarial losses, which are costly to compute and notoriously difficult to train, rely on diffusion models, or employ conventional losses such as LPIPS~\cite{LPIPS} or L1, which often fail to match the performance of adversarial supervision.
A few works explore DINOv2-based losses~\cite{ESCT3D, dino_ir, mtred, hf_diff}, or more generally rely on foundation model features for perceptual similarity~\cite{dreamsim, dreamsim2}, but they typically focus on 2D tasks or neglect intermediate representations. 
Our approach leverages intermediate, multi-layer DINOv2~\cite{dinov2} features in combination with SAM2.1~\cite{sam2} image encodings to improve 3D head consistency and quality for high-frequency areas.

\subsection{3D Face Editing}
Several methods perform 3D editing in the latent space of NeRF-based GANs~\cite{Li_2023_CVPR, Xie_2023_CVPR, ide3d, Lan_2023_CVPR, nfe}, but they do not support natural language-based edits and instead rely on style reference images or manually specified attributes. There have also been various text-guided 3D editing models introduced \cite{li2024instructpix2nerfinstructed3dportrait, 10550583, para2024avatarmmc3dheadavatar, gal2021stylegannada}, but these have core limitations like view-dependent artifacts and limited 3D consistency.

ClipFace~\cite{aneja2023clipface}, on the other hand, enables text-guided editing of 3D Morphable Models (3DMMs), but in doing so inherits the limited geometric expressiveness and visual fidelity of mesh-based representations.

LAM~\cite{lam} enables text- and image-based 3D avatar generation using diffusion models, but the editing process occurs entirely in 2D and is later lifted into 3D space. This design incurs substantial computational cost and limits direct control over 3D structure. 

In contrast, our method uses FARL segmentation maps and CLIP features as 2D encodings that disentangle geometry and style, and directly attends to them in 3D space to synthesize the full head efficiently and with explicit control.
\begin{figure*}[t]
  \centering
  \includegraphics[width=\textwidth]{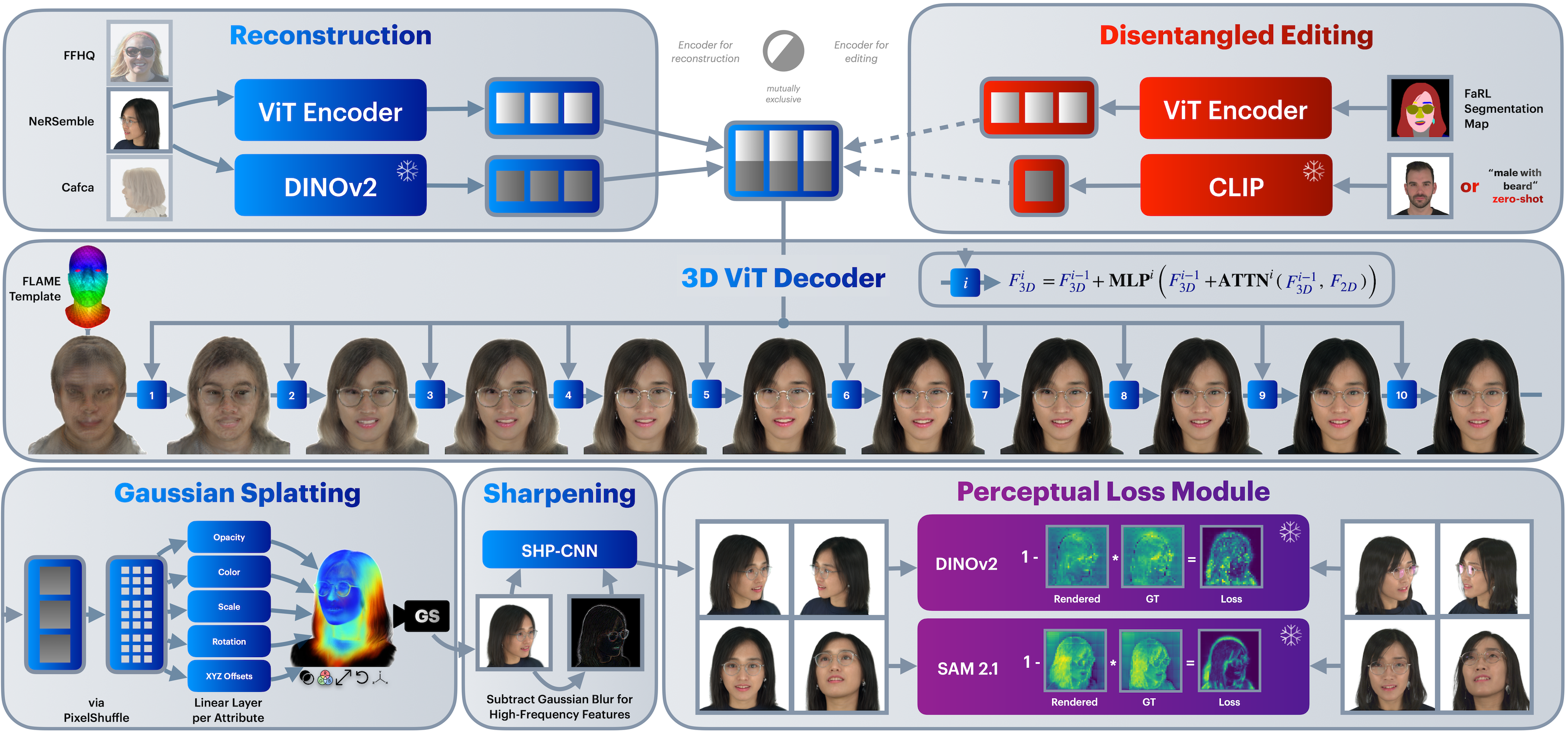}
  \caption{\textbf{Overview of Our Method.} Our framework supports \textit{3D Reconstruction} from a single image and \textit{3D Editing} from a segmentation map and style input. We only need to adapt the encoders of our ViT-based architecture between the two tasks. We train on multi-view datasets and extend them with single-view data to improve diversity, without degrading 3D consistency. 
  At the heart of our method is the supervision through DINOv2 \cite{dinov2} and SAM 2.1 \cite{sam2}. Our perceptual loss compares deep image representations, which helps us avoid signals from pixel-level differences.}
  \label{fig:method}
  \vspace{-0.5em} 
\end{figure*}

\section{Method}
\label{sec:method}

Our primary goal is to build a model which learns to generate completely 3D-consistent heads from a single image. To this cause, we choose a ViT-based architecture \cite{vit} as our base model, allowing us to cross-attend to 2D features in a latent 3D representation. Furthermore, our training objective should be designed to enable the model to reconstruct a complete 3D head, even from side-view inputs. To achieve this, we rely on a mix of single-view in-the-wild data and multi-view samples. Notably, because supervision for high-frequency regions is inherently difficult, we base our loss on foundation models.

Our second goal is to demonstrate that our approach naturally extends to downstream tasks, such as 3D editing, which we detail in \cref{sec:method_editing}.

A complete overview of our method is shown in \cref{fig:method}.

\subsection{ViT-based 3D Head Reconstruction}
\label{sec:method_3dvit}
We first modify MAE's ViT architecture \cite{mae} by implementing a dual-branch 2D encoder to extract information from the image:

\begin{equation} 
    F_\text{2D} = \mathbf{MLP}\left(\left[F^{i}_\text{ViT}, F^{i}_\text{DINOv2}\right]_{i=1}^{|P|}\right)
\end{equation}
$F_\text{DINOv2}$ are features extracted from multiple layers in DINOv2's representation \cite{dinov2} of the input, capturing high-level semantic understanding and low-level details. We augment this approach, which showed great results in LAM \cite{lam}, by introducing an additional trained ViT that provides features $F_\text{ViT}$. We include the ViT to allow our model to obtain task-specific reconstruction features. Both models provide the same number of patches $P$, which are concatenated and projected through an MLP. We only keep foreground patches.

Next, we initialize a decoding ViT from a FLAME template \cite{FLAME}. We upsample the template to 65k vertices and group neighbors into patches of 16. 
We'll call this initial representation $F^{0}_\text{3D}$. Later, we will reuse the vertex locations as anchors for the Gaussians. Now we can iteratively enrich the 3D representation:
\begin{equation}
    F^{i}_\text{3D} = F^{i-1}_\text{3D} + \mathbf{MLP}^{i}\left(F^{i-1}_\text{3D} + \mathbf{ATTN}^{i}\left(F^{i-1}_\text{3D}, F_\text{2D}\right)\right)
\end{equation}
At each decoding layer $i=1,...,10$, we use the current 3D latents to cross-attend to relevant 2D features. Notably, because each latent represents a local group, this information, together with the global context from the 2D features, is enough to reconstruct a coherent 3D head, without the use of any self-attention.

To transform the 3D patches into Gaussians, we use PixelShuffle \cite{pixelshuffle} to create 16 Gaussian latents each ($F_\text{Gaussian}$). 

For each Gaussian attribute, i.e., position, scale, rotation, opacity, and color, we use one independent linear layer. For position, we reuse the initial FLAME vertices as an anchors:
\begin{equation}
    X = X_\text{FLAME} + \mathbf{Tanh}\left(\mathbf{Linear}\left(F_\text{Gaussian}\right)  * s_\text{init}\right) * s_\text{max}
\end{equation}
$s_\text{init}$ controls movement early during training and $s_\text{max}$ controls how far the Gaussians can move from the anchor. We then use the Gaussian Splatting Rasterizer \cite{gaussian_splatting} to output reconstructed images given a camera position.

We further refine the images using a Sharpening-CNN (SHP-CNN). The input to the CNN is a blurred version (via Gaussian blur) concatenated together with high-frequency details extracted by subtracting the blurred version from the rasterized output, inspired by \cite{unsharpmask}. A multi-layer CNN with skip-connections then learns how to sharpen the blurred image by subtracting a learned residual.

\subsection{3D Consistent Training Task}
\label{sec:method_data}

With our training task we want our model to be robust to three factors: (1) consistency in its 3D representation, (2) consistency to side-facing inputs, and, (3) robustness to diverse identities. 

We'll handle (1) and (2) through multi-view datasets. The first dataset we choose is NeRSemble \cite{nersemble}. NeRSemble offers 270 real identities with 16 views per person. Additionally, we also use Cafca \cite{cafca}, which contains 1500 artificial personas, with 30 views per head. Cafca contains some camera positions which only capture the back of the head. We discard such views through face detection from GAGAvatar Track \cite{gagavatar}. Throughout training, the variety of input viewpoints encourages the model to robustly handle side-view inputs, while comparisons to three additional views will improve reconstruction performance in unseen regions to generate complete 3D heads.

To address factor (3), we incorporate a subset of images from the in-the-wild FFHQ dataset \cite{ffhq}. Although FFHQ contains 70k samples, we use only 3k, which is sufficient to make the model robust to extreme out-of-distribution reconstructions while maintaining 3D consistency. We additionally discard images with occlusions to further stabilize training.

\begin{table*}[t]
\centering
\small
\begin{tabular}{lccccc|ccccc}
\toprule
& \multicolumn{5}{c|}{\textbf{Ava-256} (Novel Views)} 
& \multicolumn{5}{c}{\textbf{Ava-256} (Extreme Views)} \\
\cmidrule(r){2-6} \cmidrule(l){7-11}
\textbf{Method} 
& PSNR ↑ & SSIM ↑ & LPIPS ↓ & DSim ↓ & ArcFace ↓ 
& PSNR ↑ & SSIM ↑ & LPIPS ↓ & DSim ↓ & ArcFace ↓ \\
\midrule
LGM               
& 10.5 & 0.591 & 0.527 & 0.185 & 0.722 
&  9.9 & 0.586 & 0.565 & 0.262 & 0.737 \\
PanoHead          
& 14.8 & 0.687 & 0.314 & 0.106 & 0.358 
& 14.2 & 0.658 & 0.349 & 0.131 & 0.416 \\
SphereHead          
& 14.7 & 0.693 & 0.317 & 0.110 & 0.379
& 14.3 & 0.661 & 0.362 & 0.149 & 0.454 \\
GAGAvatar         
& 15.9 & \textbf{0.743} & 0.274 & 0.114 & 0.348 
& 13.5 & \textbf{0.694} & 0.364 & 0.183 & 0.523 \\
LAM               
& 13.8 & 0.695 & 0.353 & 0.143 & 0.406 
& 11.4 & 0.634 & 0.455 & 0.219 & 0.568 \\
Ours              
& \textbf{16.4} & 0.691 & \textbf{0.269} & \textbf{0.092} & \textbf{0.292} 
& \textbf{15.9} & 0.678 & \textbf{0.291} & \textbf{0.106} & \textbf{0.282} \\
\bottomrule
\end{tabular}
\caption{\textbf{Quantitative results for the unseen dataset Ava-256 \cite{ava256}.} We compare performance on novel- and extreme-view reconstruction.}
\label{tab:ava256_novel_extreme}
\end{table*}

\subsection{Perceptual Loss Module}
\label{sec:method_supervision}

During training, the model is required to reconstruct large regions that are never directly observed. As a result, producing pixel-accurate outputs, particularly in high-frequency areas, is unrealistic. Under these conditions, pixel- or low-level image-space losses such as L1, SSIM \cite{SSIM}, or LPIPS \cite{LPIPS} provide noisy supervision and ultimately degrade quality. Instead, we construct our loss formulation using strong, pretrained vision models.

First, we make use of DINOv2's \cite{dinov2, dinov2reg} high-level understanding, by comparing mid to late layer activations between our rendered image $I_\text{r}$ and the ground truth $I_\text{gt}$.
\begin{equation}
\mathcal{L}_{\text{D}}
= \sum_{l\in L}\sum_{t\in T_D} \frac{
\Big(
1 - \mathrm{CD}_{l,t}(I_{\mathrm{r}})\cdot \mathrm{CD}_{l,t}(I_{\mathrm{gt}})
\Big)
}{|L|\,|T_D|}
\end{equation}
\begin{equation}
\mathrm{CD}_{l,t}(I)
= \frac{\mathbf{DINOv2}_l(I)_t}{\left\|\mathbf{DINOv2}_l(I)_t\right\|_2}
\end{equation}
With $L=\{8, 11\}$ being the intermediate layers, and $T_D$ being the class and patch tokens. Different layers can have different effects on the trained model: For example, mid layers can improve sharpness, while late layers will focus more on semantic concepts like glasses or hair. 

Further, we utilize SAM 2.1 \cite{sam2} as a segmentation model, which can supervise details correctly. Instead of comparing the segmentation maps, we use the intermediate image encodings. 

\begin{equation}
\mathcal{L}_{\text{S}}
= 
\sum_{t\in T_S}
\frac{
\left(
1 -
\frac{
\mathbf{SAM2{.}1}(I_{\text{r}})_t \cdot \mathbf{SAM2{.}1}(I_{\text{gt}})_t
}{
\left\|\mathbf{SAM2{.}1}(I_{\text{r}})_t\right\|_2\;
\left\|\mathbf{SAM2{.}1}(I_{\text{gt}})_t\right\|_2
}
\right)
}{|T_S|}
\end{equation}

These two models are enough to supervise our method without any additional losses:
\begin{equation}
    \mathcal{L} = \lambda_\text{D} \mathcal{L}_{\text{D}} *  \lambda_\text{S} \mathcal{L}_{\text{S}}
\end{equation}
With $\lambda_\text{D} = \lambda_\text{S} = 1.0$. We use this loss formulation for both the reconstruction and editing pipelines. The only exception is the SHP-CNN, which is supervised using the 8th-layer class token from DINOv2. Empirically, this yields the best sharpness. For both foundation models, we rely on the distilled small variants, resulting in a combined parameter count of 67M.

\subsection{3D Editing Model}
\label{sec:method_editing}

To adapt our reconstruction model for editing, we modify only the encoder. We replace the single image input with two new inputs: The first is a segmentation map with 19 channels, which can be obtained through FaRL \cite{farl}. The second input is a single CLIP \cite{clip} embedding, which will control style. Disentanglement between the two emerges without any other modifications.
\begin{equation} 
    F_\text{Geom+Style} = \mathbf{MLP}\left(\left[F^{i}_\text{ViT-Geom}, F_\text{CLIP}\right]_{i=1}^{|P|}\right)
\end{equation}
We initialize the rest of the pipeline with reconstruction weights and fine-tune the new model. For data, we can reuse the images as well, this time excluding samples from Cafca to avoid generating artificial heads. We input the same image twice into CLIP and FFHQ. 

At inference time, we control geometry via the segmentation map, and support even hand-drawn edits through our interactive GUI. For style control, we can either use images, or CLIP's text encoder to prompt the model in a zero-shot manner. A demo of our interactive GUI can be found in our supplementary video.
\section{Results}
\label{sec:results}

\begin{figure*}[t]
  \centering
  \includegraphics[width=1.0\textwidth]{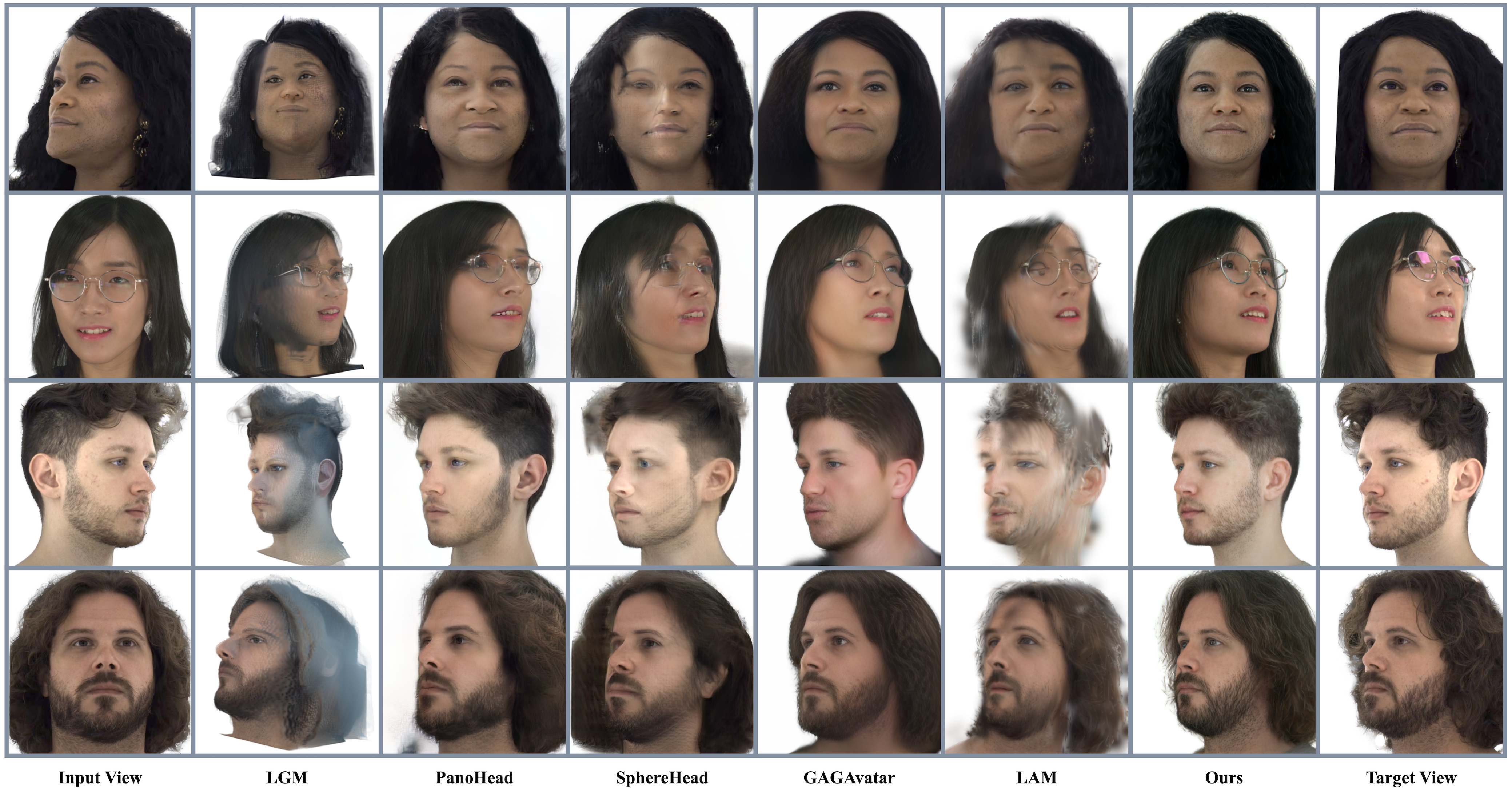}
  \caption{\textbf{Qualitative evaluation on samples from Ava-256 \cite{ava256} and NeRSemble \cite{nersemble}.} Ava-256 is a completely unseen dataset. The NeRSemble sample (person with glasses) is a held-out identity.}
  \label{fig:qual_results}
  \vspace{-0.5em} % (optional) tighten after caption if needed
\end{figure*}

\subsection{Experiment Setup}
\label{sec:results_setup}

We evaluate our model on the unseen dataset Ava-256 \cite{ava256}. We perform comparisons across two tasks: novel-view synthesis and extreme-view synthesis. For extreme views on Ava-256 we render the corresponding opposing horizontal view, which tests whether the reconstructed head is view-independent and whether the method can plausibly infer large unseen regions. We also show more evaluations in our supplementary material, including held-out identities from NeRSemble \cite{nersemble} with large vertical view shifts.

\noindent\textbf{Baselines:}\\
\noindent\textit{LGM} \cite{lgm}: Diffusion-based multi-view Gaussian model.\\
\noindent\textit{PanoHead} \cite{panohead}: 3D GAN using a tri-grid representation with a NeRF and neural upsampler; PTI inversion \cite{pti} is applied for 3D reconstruction.\\
\noindent\textit{SphereHead} \cite{spherehead}: 3D GAN employing a spherical tri-plane and a view--image consistency loss to mitigate back-view artifacts in full-head synthesis.\\
\noindent\textit{GAGAvatar} \cite{gagavatar}: Lifts Gaussians from 2D feature maps and renders them with an additional neural renderer.\\
\noindent\textit{LAM} \cite{lam}: Transformer-based Gaussian avatar generator trained on VFHQ \cite{vfhq} and NeRSemble \cite{nersemble}.\\
\noindent\textit{FaceLift} \cite{facelift}: See supplementary \cref{sec:facelift_comparison}.

\begin{figure}[t]
  \centering
  \includegraphics[width=1.0\linewidth]{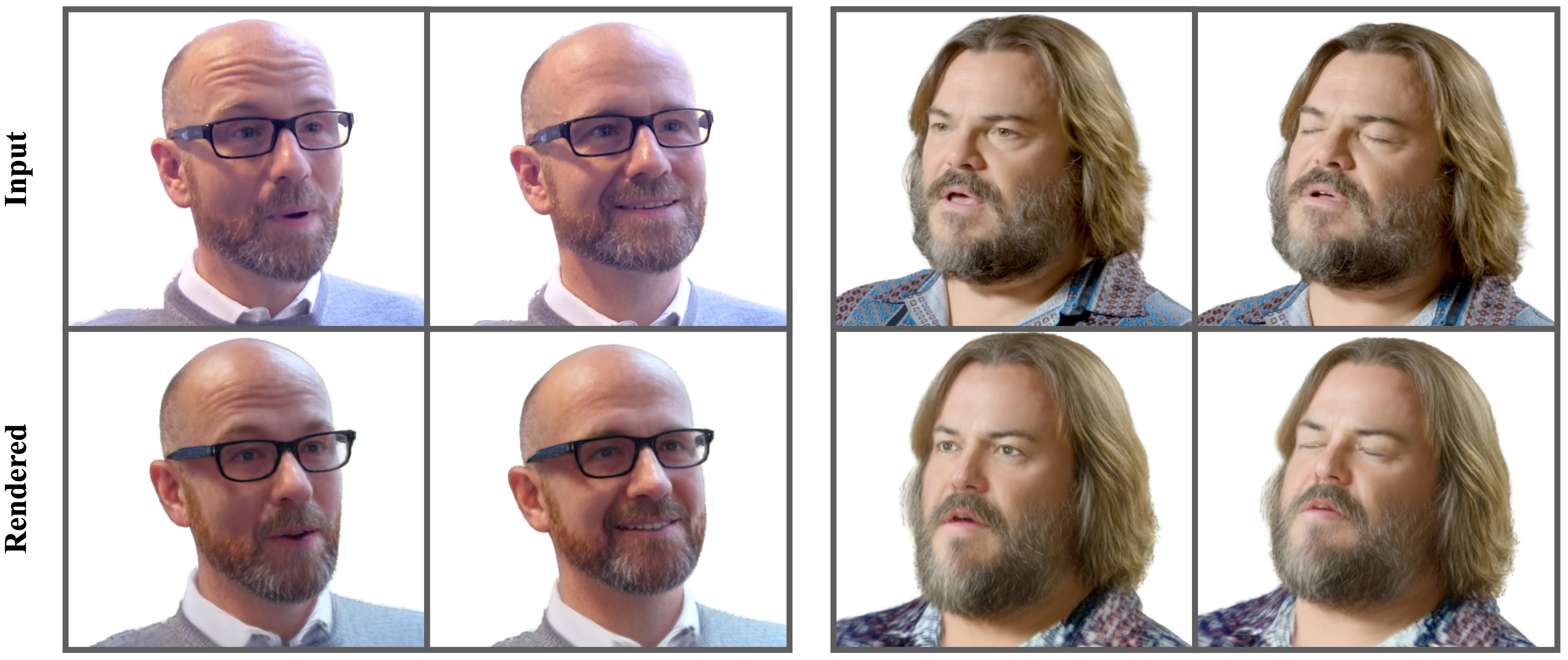}
  \caption{\textbf{3D Reconstructions Across Video Frames.} Our model maintains consistent geometry and appearance across time, enabling coherent 3D avatar lifting while capturing subtle expression changes like mouth, eye, and eyelid movements.}
  \label{fig:video_frames}
  \vspace{-0.5em} 
\end{figure}

\begin{figure*}[t]
  \centering
  \includegraphics[width=\textwidth]{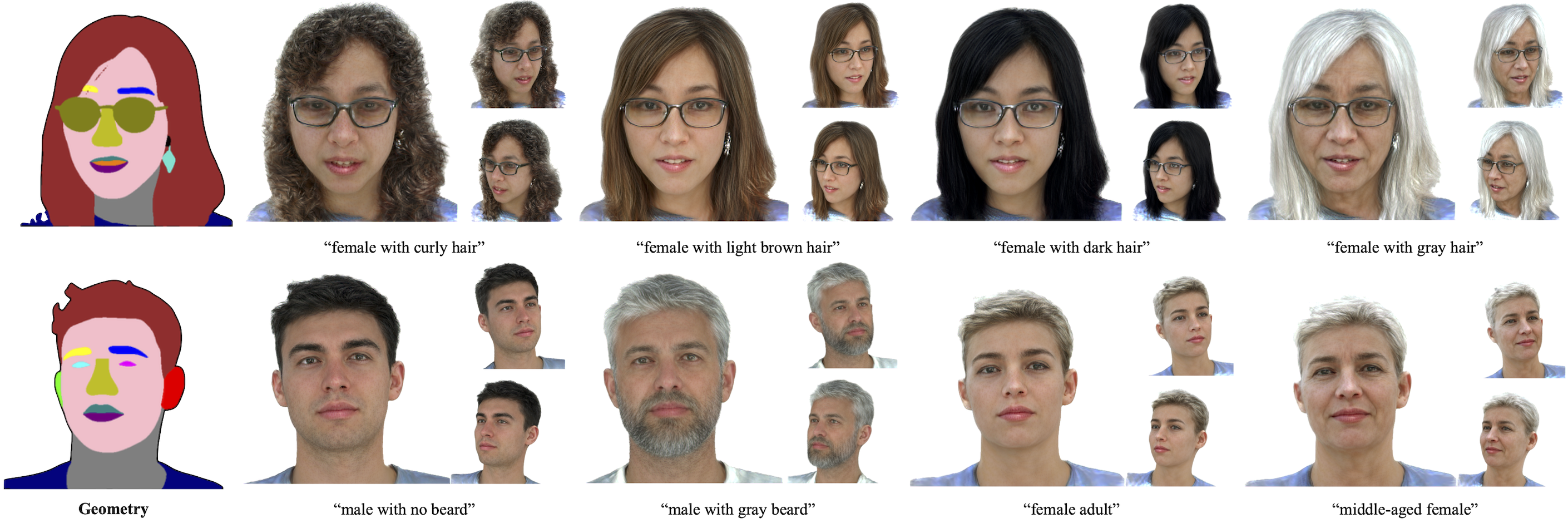}
  \caption{\textbf{Text-Based 3D Editing.} Given a fixed segmentation map and varying text prompts, our model generates diverse 3D heads with consistent geometry. Styles are guided by text, enabling low-level (e.g., hair color) and high-level (e.g., age) edits. Despite no text-specific training, our model achieves zero-shot editing via the vision-aligned CLIP \cite{clip} text encoder.}
  \label{fig:cond_results_text}
  \vspace{-0.5em} 
\end{figure*}

\begin{figure}[t]
  \centering
  \includegraphics[width=\linewidth]{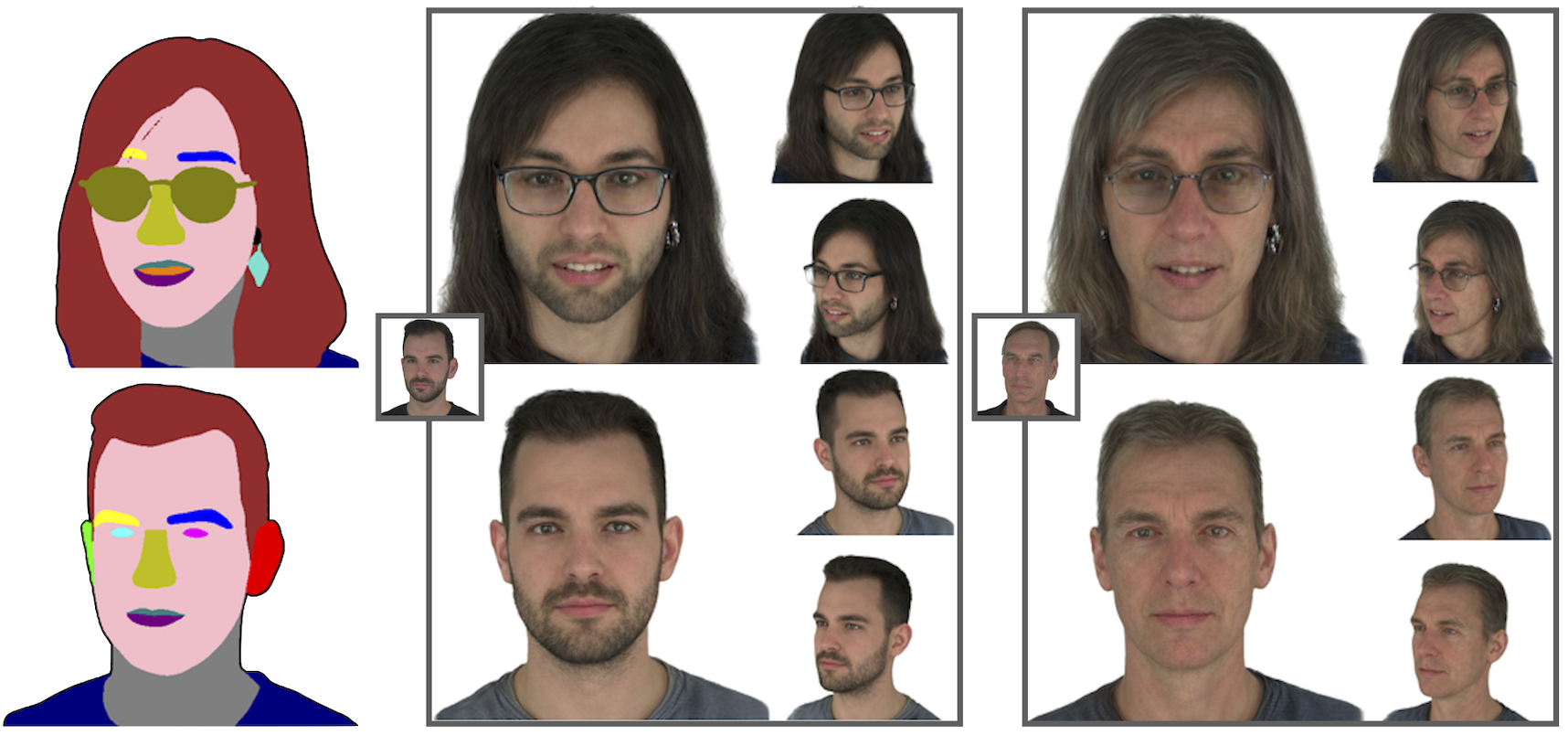}
  \caption{\textbf{Conditional 3D Head Generation from Geometry and Style.} Our method disentangles geometry and style, enabling diverse style transfer on fixed geometry and consistent appearance across varying geometries for a given style.}
  \label{fig:cond_results_img}
  \vspace{-0.5em} 
\end{figure}
\begin{figure}[t]
  \centering
  \includegraphics[width=1.0\linewidth]{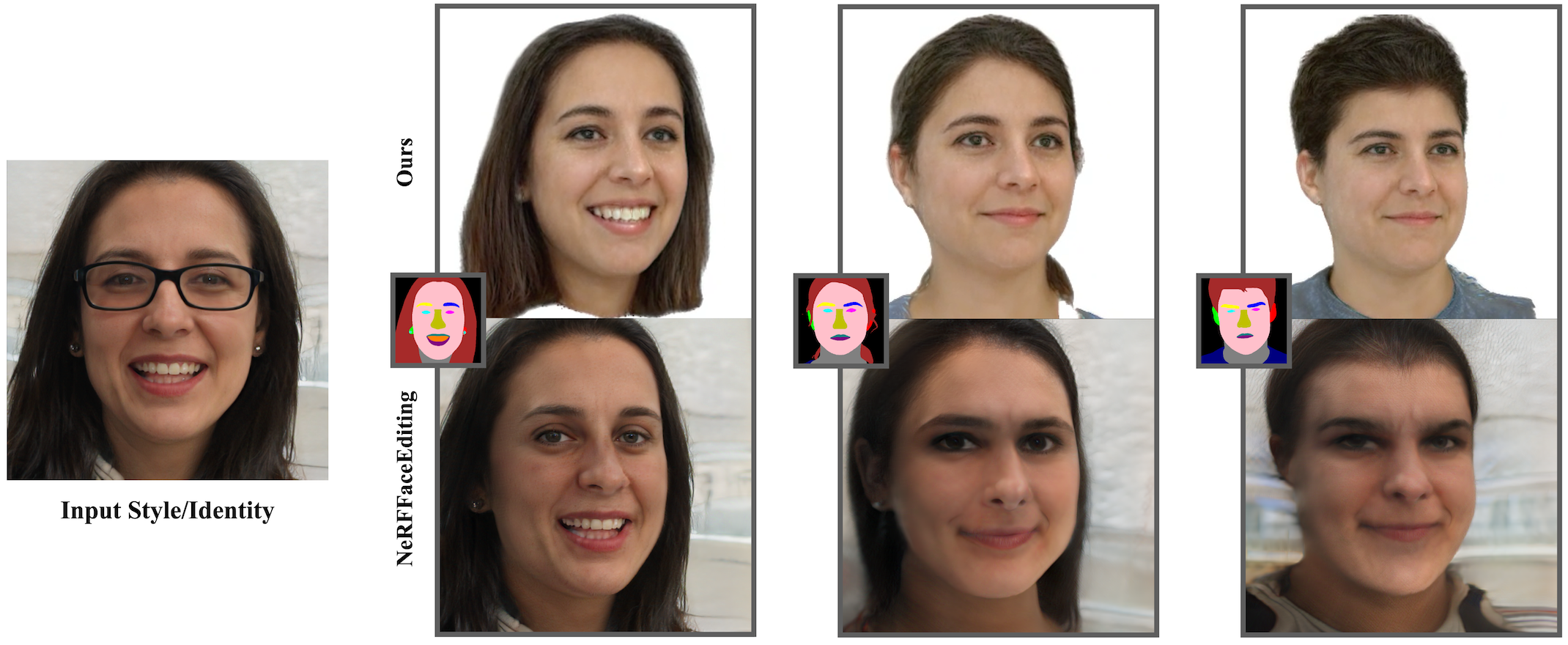}
  \caption{\textbf{We compare our geometric editing with NeRFFaceEditing (NFE) \cite{nfe} on their demo subject.} NFE performs better on small localized edits, while we perform better on large-scale edits.}
  \label{fig:nfe_comp}
  \vspace{-0.5em} 
\end{figure}

\subsection{Quantitative Results}
\label{sec:results_quantitative}

For our quantitative evaluation in \cref{tab:ava256_novel_extreme}, we report PSNR and SSIM \cite{SSIM} as standard reconstruction metrics, LPIPS \cite{LPIPS} and DreamSim (DSim) \cite{dreamsim} as perceptual metrics, and ArcFace \cite{arcface} distance to assess identity preservation.

Across both tasks, our method consistently outperforms all baselines in terms of PSNR, LPIPS, DreamSim, and ArcFace, demonstrating close low-level reconstruction, high perceptual realism, and high identity preservation. The results indicate that our model is exceptionally well-suited for 3D-consistent reconstructions. While GAGAvatar achieves slightly higher SSIM, our method achieves superior results in all other metrics, particularly those better aligned with human perceptual quality and identity preservation.

The performance gap becomes even more pronounced in the extreme views scenario. The best performing baseline, PanoHead, degrades significantly in LPIPS (+ 11\%), DreamSim (+23\%), and ArcFace (+16\%). In contrast, our method remains signifcantly more robust under these challenging conditions, even showing slightly better scores in ArcFace (+8\%, +15\%, -3\%).

\subsection{Qualitative Results}
\label{sec:results_qualitative}

\cref{fig:qual_results} presents a visual comparison between methods under both novel-view and extreme-view conditions. PanoHead frequently exhibits mirroring artifacts, generating implausible completions in occluded areas. SphereHead also exhibits clear 3D-consistency failures, frequently mis-reconstructing glasses, hair, and even core facial geometry across viewpoints. GAGAvatar struggles with frontal-to-side transitions, losing detail and geometric fidelity, and often collapses under more extreme conditions. LAM fails to maintain 3D consistency in its predicted Gaussians, leading to structural distortions when rendered from challenging views. LGM also degrades in novel views, often hallucinating unrealistic head geometry and adding a blue tint. In contrast, our model consistently produces realistic and 3D-consistent reconstructions, maintaining detail and coherence even under wide viewpoint changes. It accurately completes unseen regions without introducing artifacts or degrading identity, highlighting its strong generalization and geometric understanding.

Another common downstream task is generating 3D avatars from video. Although our model is trained on single-frame images, we show that it generalizes to this setting by applying it frame-by-frame to video input. Using two sequences from the unseen dataset VFHQ \cite{vfhq}, we evaluate our model's ability to handle changes in facial expressions across time. As shown in \cref{fig:video_frames}, our model produces reconstructions that are faithful to each frame while remaining consistent in geometry and appearance. However, we do want to mention that our model will introduce slight flickering for unobserved regions. This is only noticeable in the video sequences, which we include in our supplementary video.

\begin{figure*}[t]
  \centering
  \includegraphics[width=\textwidth]{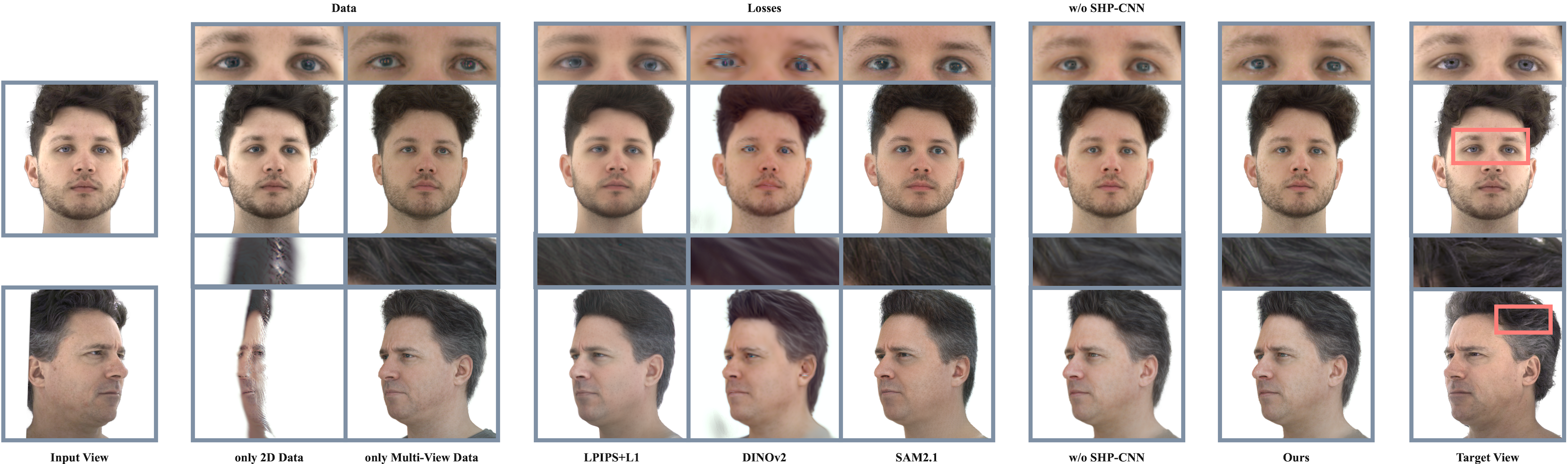}
  \caption{\textbf{Ablation Study on Data and Loss Variants.} We compare 3D head reconstruction results for models trained with: 2D data only, multi-view data only, LPIPS + L1 loss, DINOv2 loss, SAM2.1 loss, without the sharpening module, and our full configuration.}
  \label{fig:ablations}
  \vspace{-0.5em} % (optional) tighten after caption if needed
\end{figure*}

\subsection{Disentangled 3D Editing}
\label{sec:disentangled_3d_editing}

As shown in \cref{fig:cond_results_img}, our framework can generate 3D heads by combining structural guidance from a segmentation map with style cues extracted from a reference image. The model preserves the input geometry when applying diverse styles, and conversely, maintains coherence when transferring the same style across different geometries. We also compare our method visually against NeRFFaceEditing (NFE) \cite{nfe} in \cref{fig:nfe_comp}. For localized edits, NFE performs extremely well, but breaks for large-scale edits. Our model is limited through CLIP's \cite{clip} single input embedding, offering only limited identity preservation. However, we observe that the reconstructed identity stays consistent throughout all edits.

Building on this capability, \cref{fig:cond_results_text} demonstrates that our model also supports text-driven 3D head editing in a zero-shot manner. While it is trained only on image-based style conditioning, we leverage the vision-aligned CLIP text encoder to map prompts into the same latent space used for style control. This allows our approach to interpret and apply textual edits without any task-specific fine-tuning. The model responds robustly to both low-level visual edits, such as altering hair color or texture (e.g., curly hair), and high-level attributes, including age, where it adapts skin and hair characteristics accordingly. We also incldue a demo of our interactive GUI for this task in our supplementary video.

\subsection{Ablations}
\label{sec:ablations}

\begin{table}[t]
\vspace{0.0em}
\centering
\scriptsize
\renewcommand{\arraystretch}{0.95}
\setlength{\tabcolsep}{4pt}
\begin{tabular}{@{}llcccc@{}}
\toprule

 & Variant & PSNR ↑ & LPIPS ↓ & DreamSim ↓ & ArcFace ↓ \\
\midrule
\multirow{3}{*}{\rotatebox[origin=c]{90}{\emph{Data}}}
& 2D Data                 & 6.5  & 0.643 & 0.436 & 0.752 \\
& Multi-View Data         & 15.6 & 0.299 & 0.111 & 0.308 \\
& 2D + MV data (Ours)     & \textbf{15.9} & \textbf{0.291} & \textbf{0.106} & \textbf{0.282} \\
\cmidrule(lr){1-6}
\multirow{4}{*}{\rotatebox[origin=c]{90}{\emph{Loss}}}
& LPIPS+L1                & \textbf{15.9} & \textbf{0.285} & 0.115 & 0.300 \\
& DINOv2                  & 15.4 & 0.383 & 0.149 & 0.340 \\
& SAM2.1                  & \textbf{15.9} & 0.309 & 0.107 & 0.311 \\
& DINOv2+SAM2.1 (Ours)    & \textbf{15.9} & 0.291 & \textbf{0.106} & \textbf{0.282} \\
\cmidrule(lr){1-6}
\multirow{2}{*}{\rotatebox[origin=c]{90}{\emph{SHP}}}
& w/o SHP-CNN         & \textbf{15.9} & 0.301 & 0.116 & \textbf{0.282} \\
& with SHP-CNN (Ours)  & \textbf{15.9} & \textbf{0.291} & \textbf{0.106} & \textbf{0.282} \\
\bottomrule
\end{tabular}
\caption{\textbf{Ablation Studies on Extreme Views from Ava-256.}}
\label{tab:ablations}
\end{table}

Ablations are reported in \cref{tab:ablations} and \cref{fig:ablations}. 

\noindent\textbf{Training data.} While the model trained only on 2D data FFHQ \cite{ffhq} performs extremely well when reconstructing the same view, it does not reconstruct 3D heads. The model trained solely on mutli-view data produces good detail, but struggles in preserving the input lighting condition, due to limited variability in the controlled environments of both 3D datasets. Quantitatively, both models perform worse than the model trained on both 2D and multi-view data. 

\noindent\textbf{Losses.} We retrain a model based on standard losses like LPIPS \cite{LPIPS} and L1, together with it's own SHP-CNN. We observe that quantitatively, the model produces better LPIPS scores (-2\%), which is to be expected as we don't supervise our model with this metric. In contrast, for DreamSim and ArcFace, our model produces significantly better resutls (-8\%, -6\%), indicating better perceptual quality and identity preservation. Qualitatively, reconstructions from LPIPS+L1 are extremely smooth, leading to loss of detail in hair and facial features.

The DINOv2-supervised model provides results that are noticeably blurry, while producing better hair fidelity than the other two losses. DINOv2's deep vision features provide good supervision for high-frequency areas, while lacking on low-level aspects, such as sharpness, which leads to worse scores across all metrics.

A SAM 2.1-based loss leads to a model that provides great facial reconstruction, but short, cut-off hair strands, indicating that details can be differentiated well, but that hair is too noisy to be effectively supervised by segmentation-oriented encodings. Compared to our full method's results, performance remains similar for PSNR and DreamSim, but is worse for LPIPS and ArcFace.

We offer an additional analysis on our perceptual loss in our supplementary material.

\noindent\textbf{SHP-CNN.} Both qualitatively and quantitatively, our CNN only improves sharpness. It improves perceptual quality measured by LPIPS and DreamSim, but has no effect on PSNR or identity (ArcFace).

\subsection{Limitations}
Our method also has limitations: (1) It lacks support for expression transfer, which is essential for tasks like avatar animation. (2) It doesn't have any built-in temporal understanding, leading to flickering in video generation. (3) It also has baked in lighting, which limits transfer to new environments after reconstruction.
\section{Conclusion}
\label{sec:conclusions}
We presented a new state-of-the-art approach for view-angle robust 3D head reconstruction from a single image. We introduce a novel and surprisingly effective loss formulation that relies on the perceptual understanding of foundational models and demonstrate that it can be used to train for high 3D consistency and detail. This challenges the prevailing assumption that conventional image losses (e.g., L1, LPIPS \cite{LPIPS}, SSIM \cite{SSIM}) are necessary for supervision.

In addition, our editing model enables a disentangled editing interface where geometry is guided by segmentation maps and style by text or reference images, opening new possibilities for intuitive 3D face editing.
\section*{Acknowledgements}
\label{sec:acknowledgements}
Our work was supported by the ERC Consolidator Grant Gen3D (101171131). Additionally, we thank Angela Dai for the voice-over of our video. 

{
    \small
    \bibliographystyle{ieeenat_fullname}
    \bibliography{main}
}

% WARNING: do not forget to delete the supplementary pages from your submission 
\clearpage
\setcounter{page}{1}
\maketitlesupplementary

\appendix

\begin{table*}[t]
\centering
\small
\begin{tabular}{lccccc|ccccc}
\toprule
\multirow{2}{*}{\shortstack[l]{\rule{0pt}{1.2em}Method}} 
& \multicolumn{5}{c|}{\textbf{NeRSemble} (Novel Views)} 
& \multicolumn{5}{c}{\textbf{NeRSemble} (Extreme Views)} \\
\cmidrule(r){2-6} \cmidrule(l){7-11}
& PSNR ↑ & SSIM ↑ & LPIPS ↓ & DS ↓ & ArcFace ↓ 
& PSNR ↑ & SSIM ↑ & LPIPS ↓ & DS ↓ & ArcFace ↓ \\
\midrule
LGM               & 11.4 & 0.662 & 0.500 & 0.189 & 0.701 &  9.9 & 0.641 & 0.563 & 0.243 & 0.758 \\
PanoHead          & 15.4 & 0.750 & 0.261 & 0.106 & 0.338 & 14.4 & 0.716 & 0.310 & 0.119 & 0.373 \\
SphereHead        & 15.3 & 0.758 & 0.260 & 0.106 & 0.336 & 14.3 & 0.720 & 0.321 & 0.135 & 0.433 \\
GAGAvatar         & 16.9 & \textbf{0.798} & 0.217 & 0.107 & 0.288 & 14.6 & \textbf{0.750} & 0.317 & 0.162 & 0.476 \\
LAM               & 14.5 & 0.758 & 0.303 & 0.134 & 0.353 & 11.6 & 0.682 & 0.433 & 0.204 & 0.551 \\
Ours              & \textbf{18.4} & 0.764 & \textbf{0.190} & \textbf{0.071} & \textbf{0.256} 
                  & \textbf{17.7} & 0.741 & \textbf{0.221} & \textbf{0.081} & \textbf{0.257} \\
\bottomrule
\end{tabular}
\caption{\textbf{Novel and Extreme View Reconstruction Performance on NeRSemble (Unseen Identities) \cite{nersemble}.}}
\label{tab:nersemble_novel_extreme}
\end{table*}

\section{Supplementary Video}
We highly recommend watching our supplementary video, which showcases additional 3D reconstruction orbit views, frame-by-frame 3D video generation, 3D edit orbit sequences, and a live demo of our interactive 3D editing web application.

\section{Perceptual Losses During Early Training}
\begin{figure*}[t]
  \centering
  \includegraphics[width=\textwidth]{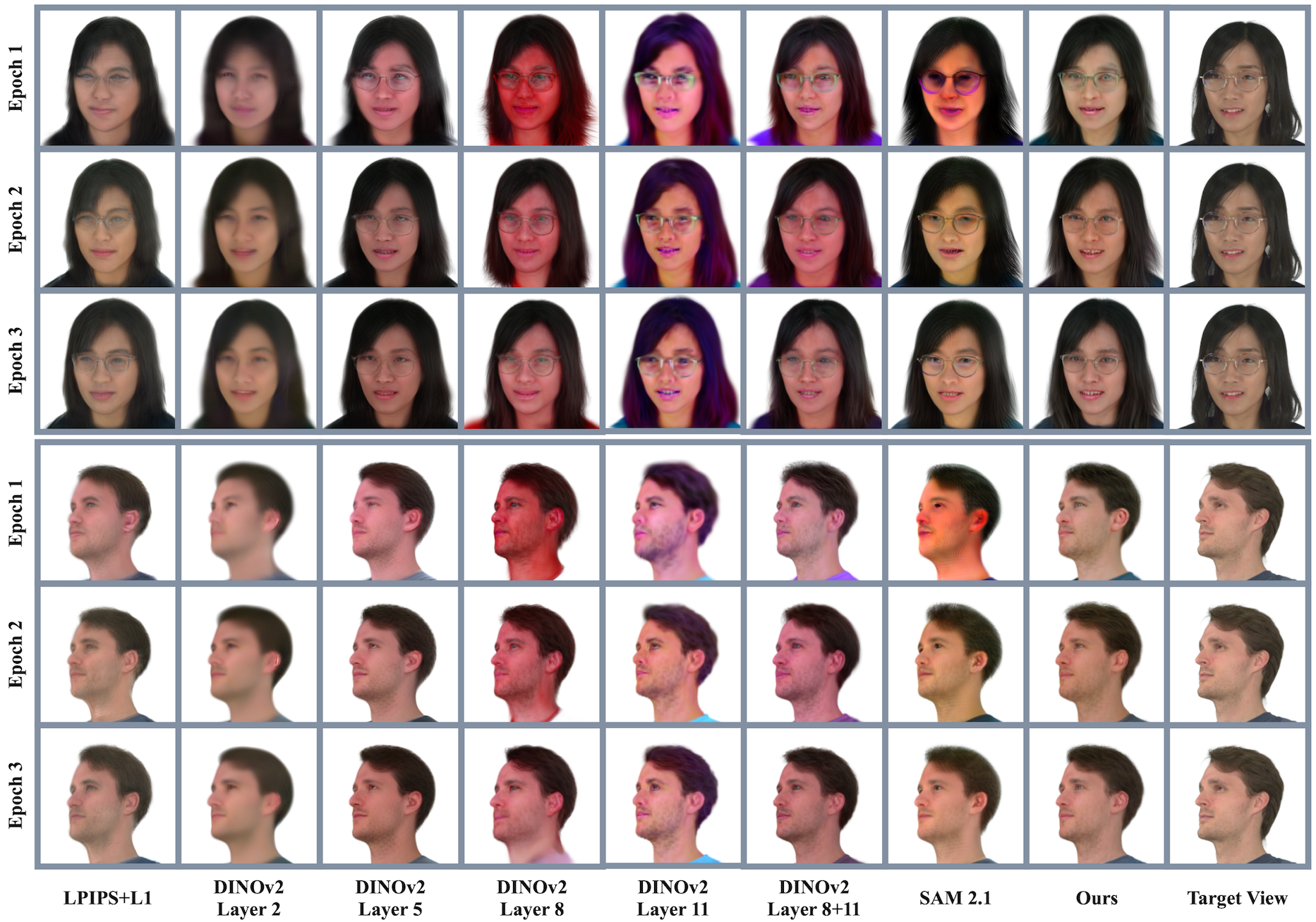}
  \caption{\textbf{Early-Epoch Comparison of Perceptual Losses.} We visualize epochs 1--3 for LPIPS+L1, different DINOv2 layers, SAM2.1, and our full model. DINOv2+SAM2.1 provides the best supervision, recovering glasses, hair, and teeth structures much earlier and with fewer color artifacts than the alternatives.}
  \label{fig:x_abl_layers}
\end{figure*}

In \cref{fig:x_abl_layers}, we compare early-epoch behavior (epochs 1--3) of different perceptual losses: LPIPS+L1, individual DINOv2 layers (2, 5, 8, 11), DINOv2 (8+11), SAM2.1, and our full combined loss. LPIPS+L1 struggles to reconstruct structures and high-frequency areas: glasses of subject 1 remain vague and incomplete even at epoch 3, teeth in subject 1 are not visible at all, and hair in subject 2 has poor texture across all three epochs. In contrast, DINOv2-based losses generally reconstruct glasses much earlier, often already at epoch 1 and more clearly thereafter. DINOv2 layer 2 is overly low-level, producing blurry outputs and without any glasses or hair structure; layers 5, 8, and 11 (and their 8+11 combination) capture glasses, teeth, and hair structure more reliably. However, these layers also introduce a red tint in the reconstructions. We also note that layer-5-supervised reconstructions are quite sharp, but we still opted for using later layers in our final loss formulation, as they offer more realistic hair. These layers are a good addition to the supervision of SAM2.1, which reconstructs details, including glasses and teeth, very well but is weaker on hair and introduces a yellowish tint. The combined DINOv2 (8+11) and SAM2.1 supervision yields: strong hair texture, robust glasses and teeth reconstruction, and without the tints observed when using either model alone.

\begin{figure}[t]
  \centering
  \includegraphics[width=1.0\linewidth]{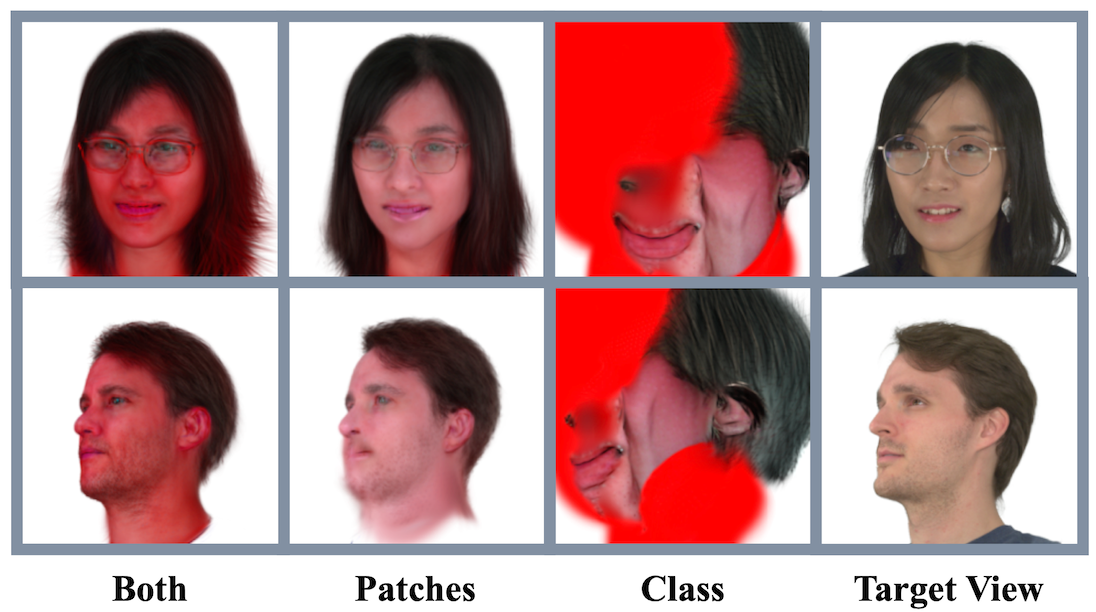}
  \caption{\textbf{Patch vs.\ Class Token Supervision (DINOv2 Layer 8).} After one epoch, patch-only supervision yields coherent but blurrier geometry, while class-only supervision produces sharper yet geometrically unstable reconstructions. Combining patch and class information offers best guidance.}
  \label{fig:x_abl_class_patch}
  \vspace{-1.0em}
\end{figure}

In \cref{fig:x_abl_class_patch}, we further analyze DINOv2 layer~8 supervision by comparing patch-only, class-only, and the combined patch+class setup after a single epoch. Patch-only supervision produces globally coherent geometry but also noticeably blurry areas (e.g., the chin of subject~2). Supervision based on the class token alone yields some sharper, stylized textures, consistent with our observations when training the SHP-CNN, but lacks geometric guidance: since the class token provides no spatially localized information, reconstructions become unstable, lose global 3D consistency, and exhibit color bleeding (e.g., red regions spreading across the face). As a result, class-only supervision is not enough to train the full 3D pipeline from scratch. Combining patch and class information provides localized geometric signals together with sharper textures, and is therefore used in our final configuration.

\section{Additional Reconstruction Results}

\subsection{Comparison with FaceLift}
\label{sec:facelift_comparison}
\begin{figure*}[t]
  \centering
  \includegraphics[width=\textwidth]{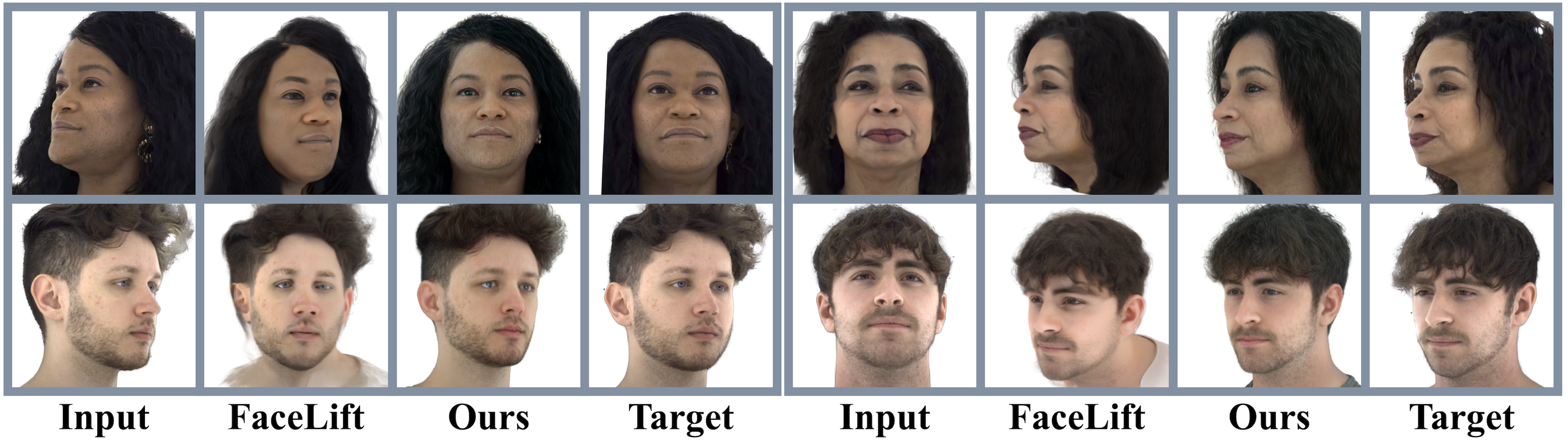}
  \caption{\textbf{Qualitative comparison with FaceLift \cite{facelift} on Ava-256 \cite{ava256}.}}
  \label{fig:x_rebuttal_facelift}
\end{figure*}

FaceLift \cite{facelift} is a single-image 3D face reconstruction method, which employs a diffusion-based multi-view generator trained on mutli-view synthetic head images. In \cref{tab:facelift_comp} and \cref{fig:x_rebuttal_facelift}, we compare our method against FaceLift. Quantitatively, our approach outperforms

Our approach outperforms FaceLift both quantitatively and qualitatively. In particular, our approach handles non-frontal inputs significantly better than FaceLift, which will often produce distorted reconstructions.
\begin{table}[h]
\centering
\scriptsize
\setlength{\tabcolsep}{2pt}
\renewcommand{\arraystretch}{1.05}
\begin{tabular*}{\linewidth}{@{\extracolsep{\fill}}lccccc}
\toprule
\textbf{Ava-256 (Novel Views)} & PSNR$\uparrow$ & SSIM$\uparrow$ & LPIPS$\downarrow$ & DSim$\downarrow$ & ArcFace$\downarrow$ \\
\midrule
FaceLift & 12.5 & 0.645 & 0.436 & 0.129 & 0.547 \\
Ours     & \textbf{16.4} & \textbf{0.691} & \textbf{0.269} & \textbf{0.092} & \textbf{0.292} \\
\bottomrule
\toprule
\textbf{Ava-256 (Extreme Views)} & PSNR$\uparrow$ & SSIM$\uparrow$ & LPIPS$\downarrow$ & DSim$\downarrow$ & ArcFace$\downarrow$ \\
\midrule
FaceLift & 11.0 & 0.627 & 0.482 & 0.181 & 0.556 \\
Ours     & \textbf{15.9} & \textbf{0.678} & \textbf{0.291} & \textbf{0.106} & \textbf{0.282} \\
\bottomrule
\end{tabular*}
\caption{\textbf{Comparison against FaceLift \cite{facelift} on Ava-256 \cite{ava256}.}}
\label{tab:facelift_comp}
\end{table}

\subsection{Quantitative NeRSemble Results}
We report quantitative results on held-out NeRSemble \cite{nersemble} identities in \cref{tab:nersemble_novel_extreme} for both the novel-views and extreme-views tasks. For the extreme setting, we additionally consider strong vertical viewpoint changes, (i.e., bottom-input-to-top-target or vice-versa). The performance mirrors our observations on Ava-256: Except for SSIM, our method consistently outperforms baselines in perceptual, identity and pixel-wise metrics.

\subsection{Qualitative Results on Ava-256 and NeRSemble}
\begin{figure*}[t]
  \centering
  \includegraphics[width=\textwidth]{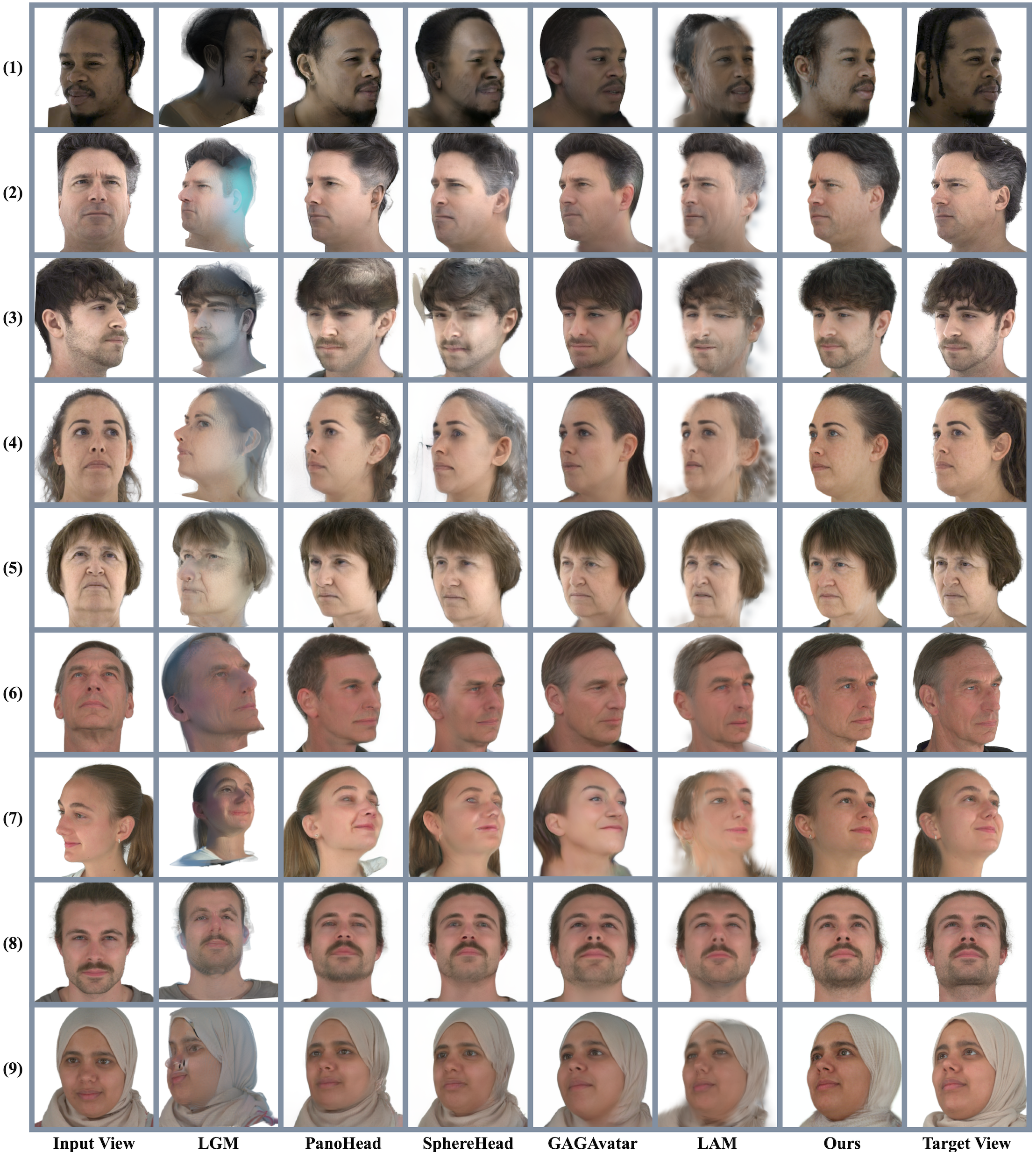}
  \caption{\textbf{Additional Qualitative Results on Ava-256 \cite{ava256} and NeRSemble \cite{nersemble}.} We show additional cross-view reconstructions on held-out subjects. Samples (1)–(5) are from Ava-256, while samples (6)–(9) are from left-out NeRSemble identities.}
  \label{fig:x_qual_ava_ners}
\end{figure*}

In \cref{fig:x_qual_ava_ners}, we present additional qualitative results on Ava-256 \cite{ava256} and NeRSemble \cite{nersemble}. The first five samples correspond to Ava-256 subjects, and the last four to left-out NeRSemble identities. Across all cases and view transitions, our method reconstructs significantly more consistent heads than competing methods, with fewer artifacts around challenging regions such as hair, ears, and jawline, and with improved identity and shape preservation under large viewpoint changes.

\subsection{Qualitative Comparison Between Target Views}
\begin{figure*}[t]
  \centering
  \includegraphics[width=0.85\textwidth]{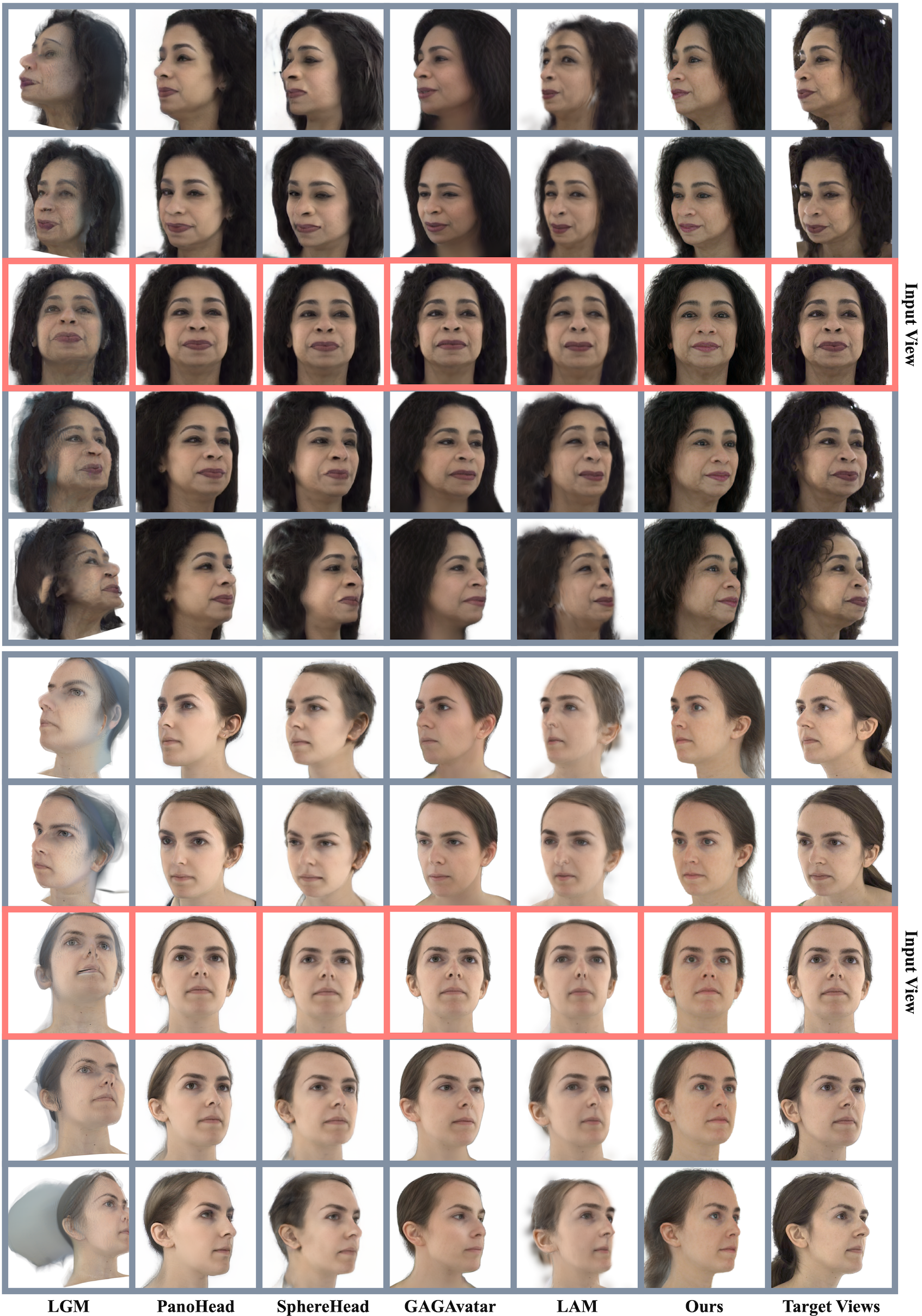}
  \caption{\textbf{Multi-View Reconstruction from a Frontal Input on Two Samples from Ava-256 \cite{ava256}.} Starting from a single frontal input view, we reconstruct a wide range of target viewpoints on a NeRSemble subject. While most methods typically perform well on the frontal view but degrade under large view changes, our method maintains consistent identity, facial geometry, and hair structure across all target angles. This demonstrates that our model genuinely recovers a stable 3D representation rather than overfitting to the input view.}
  \label{fig:x_qual_angles_front}
\end{figure*}

\begin{figure*}[t]
  \centering
  \includegraphics[width=0.85\textwidth]{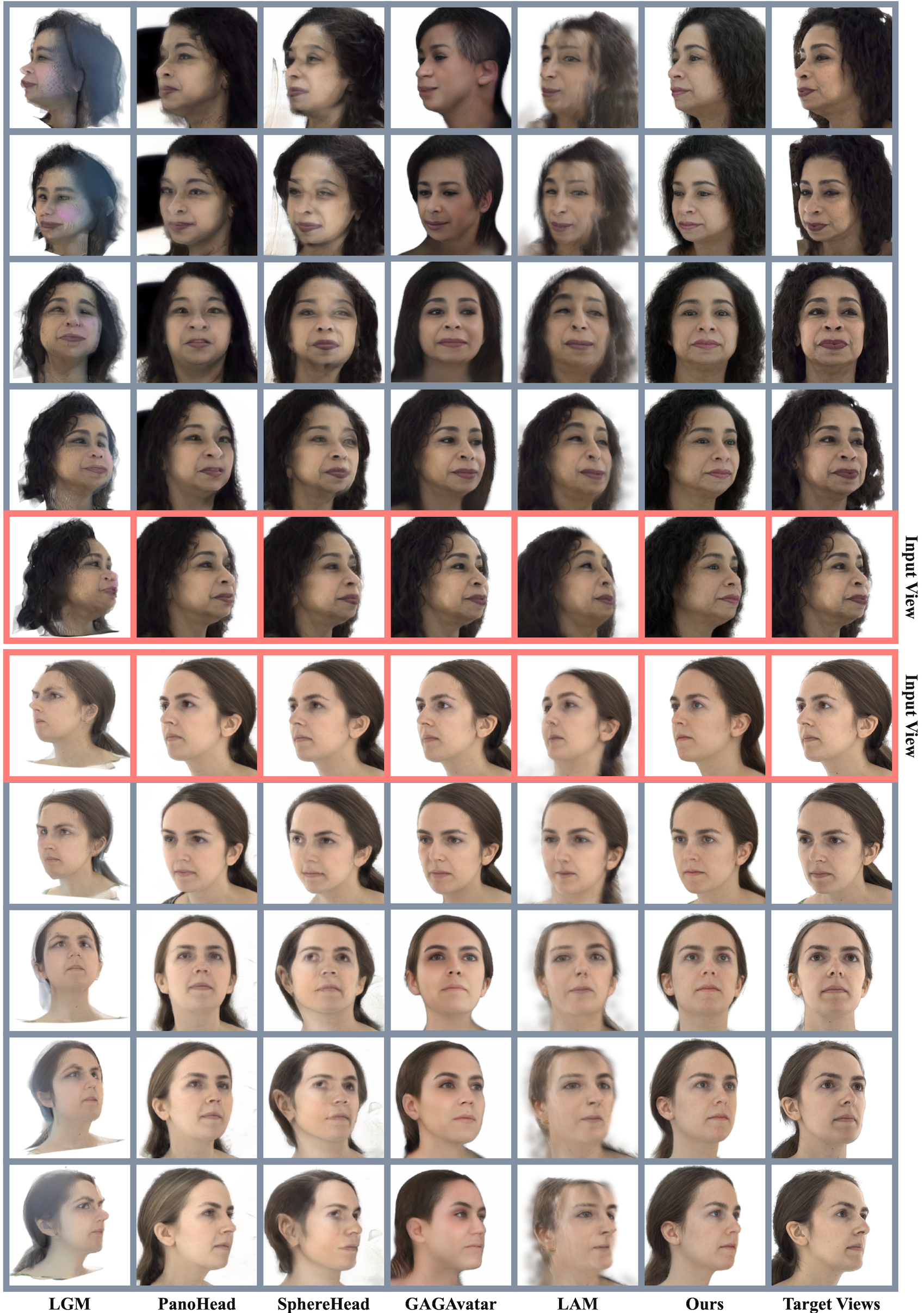}
  \caption{\textbf{Multi-View Reconstruction from a Side-View Input on Two Samples from Ava-256 \cite{ava256}.} We repeat the multi-angle reconstruction experiment using a challenging side-view input. Baseline methods often struggle to generate plausible frontal and opposite-side views, leading to identity changes, distorted shapes, or collapsed geometry. In contrast, our method produces coherent reconstructions for the input side view as well as unseen target angles, indicating strong 3D consistency and robustness in self-occluded facial regions.}
  \label{fig:x_qual_angles_side}
\end{figure*}

\Cref{fig:x_qual_angles_front,fig:x_qual_angles_side} analyze the behavior of each method when varying the target viewpoint for a fixed input image from Ava-256 \cite{ava256}. 

In \cref{fig:x_qual_angles_front}, we condition on a frontal view and render a sequence of target angles. Most baselines provide strong frontal reconstructions but quickly deteriorate when rotating away from the input view, revealing limitations in their underlying 3D consistency. Our method maintains stable identity and geometry across the full range of target angles. 

In \cref{fig:x_qual_angles_side}, we use a side-view input. Here, some baselines already struggle on the input view and introduce noticeable artifacts when extrapolating to frontal and opposite-side views. Our method remains robust for both the input and unseen angles.

\subsection{Robustness Analysis}
\subsubsection*{In-The-Wild Scenarios and Generated Inputs}
\begin{figure*}[t]
  \centering
  \includegraphics[width=\textwidth]{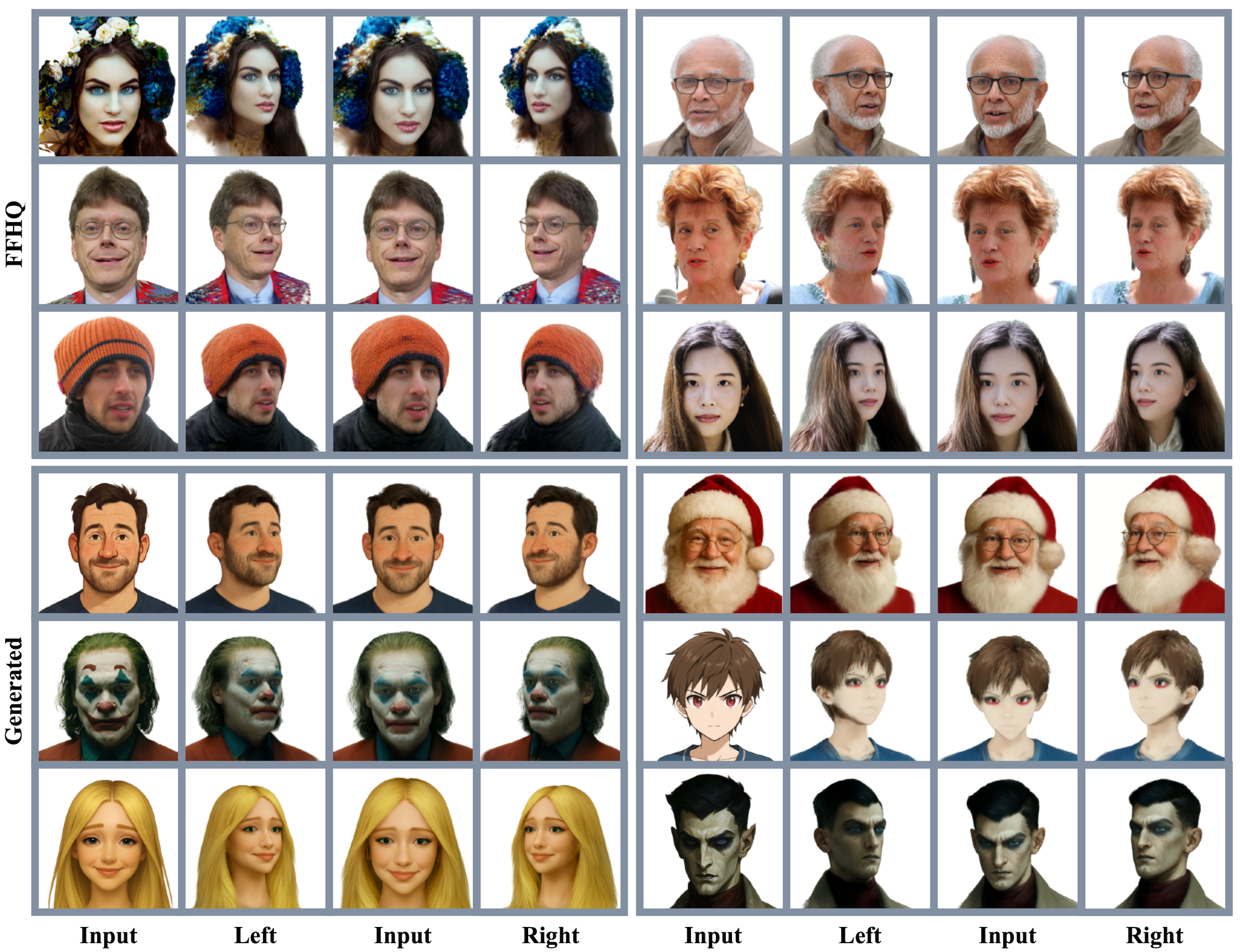}
  \caption{\textbf{Robustness on In-the-Wild and Generated Images.} We evaluate robustness on challenging inputs beyond the training distribution. The first block shows left-out FFHQ \cite{ffhq} samples, while the second block shows reconstructions of images synthesized by a diffusion model (SORA \cite{sora}), including cartoon-style, highly stylized, or otherwise complex appearances. Across both categories, our method offers extremely robust performance. Due to the training data, it does however lean towards more realistic reconstructions, even with cartoonish inputs. We also see that heavily unrealistic faces can sometimes reduce reconstruction performance - like the right sample on the second-to-last row.}
  \label{fig:x_qual_ffhq_gen}
\end{figure*}

\Cref{fig:x_qual_ffhq_gen} investigates robustness to in-the-wild and generated images. We evaluate our model on left-out FFHQ \cite{ffhq} samples and diffusion-generated images with cartoon-like, stylized, or otherwise complex appearances. Despite substantial domain shifts in pose, lighting, and style, our method continues to produce plausible 3D heads and preserves identity. These results suggest that the learned representation generalizes beyond the training distribution and remains applicable in realistic and creative downstream scenarios.

\begin{figure*}[t]
  \centering
  \includegraphics[width=\textwidth]{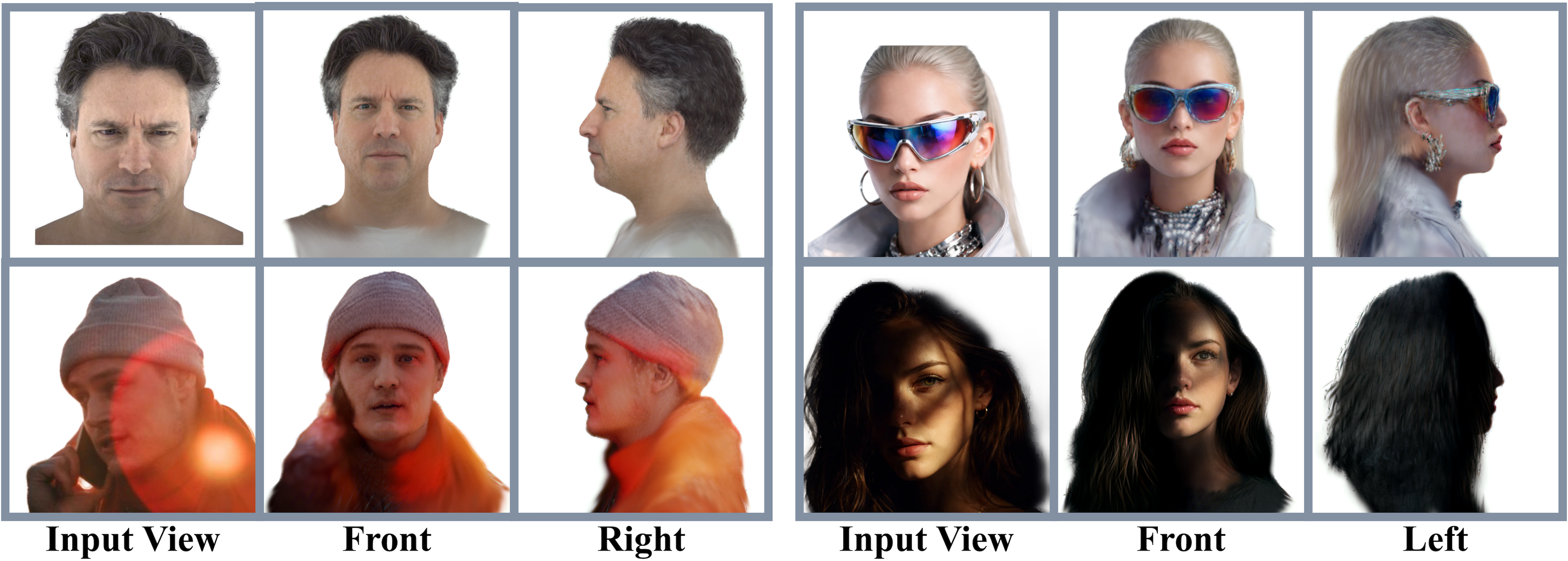}
  \caption{\textbf{Extreme Scenarios and Side Views.} Inputs are from Ava256 \cite{ava256}, SORA \cite{sora}, and UFDD \cite{ufdd}.}
  \label{fig:x_rebuttal_extreme}
\end{figure*}

In \cref{fig:x_rebuttal_extreme}, we further demonstrate that our method can produce complete side views and retains good performance even in difficult out-of-distribution scenarios with strong shadows and difficult illumination. However, we do observe that difficult geometry, such as exotic glasses, are not perfectly reconstructed by our model.
\subsubsection*{SHP-CNN in Out-Of-Distribution Scenarios}
\begin{figure*}[t]
  \centering
  \includegraphics[width=\textwidth]{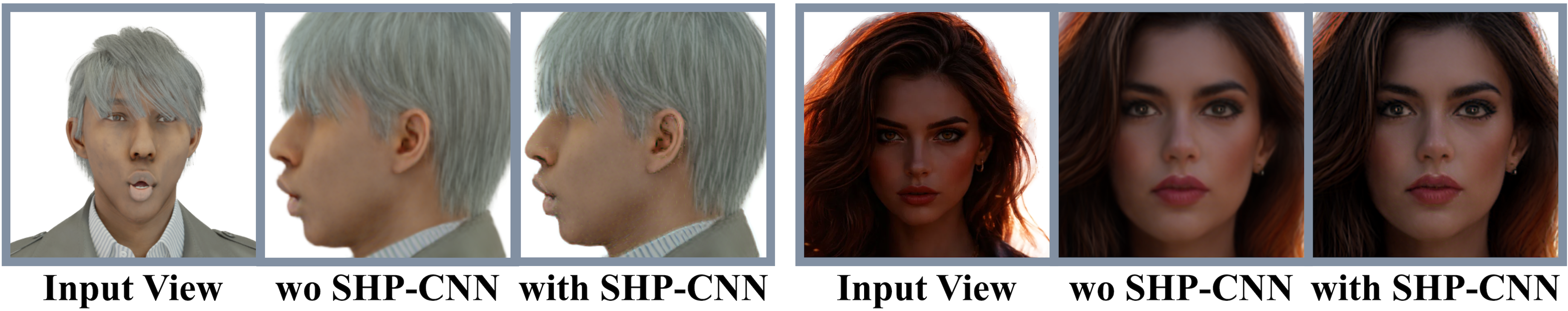}
  \caption{\textbf{SHP-CNN effects for out-of-distribution sample and difficult lighting.} Inputs are from Cafca \cite{cafca} and SORA \cite{sora}.}
  \label{fig:x_rebuttal_shp_cnn}
\end{figure*}

In \cref{fig:x_rebuttal_shp_cnn}, we that our SHP-CNN, which was only trained on Nersemble \cite{nersemble} and VFHQ \cite{vfhq} samples, also handles synthetic Cafca \cite{cafca} samples well. We also show an in-the-wild image with difficult lighting. Our SHP-CNN improves perceptual fidelity primarily by adding sharpness and stronger contrast. However, it does not add any new details. These are generated solely by the reconstruction model.

\subsubsection*{Loss Robustness}
\begin{figure*}[t]
  \centering
  \includegraphics[width=\textwidth]{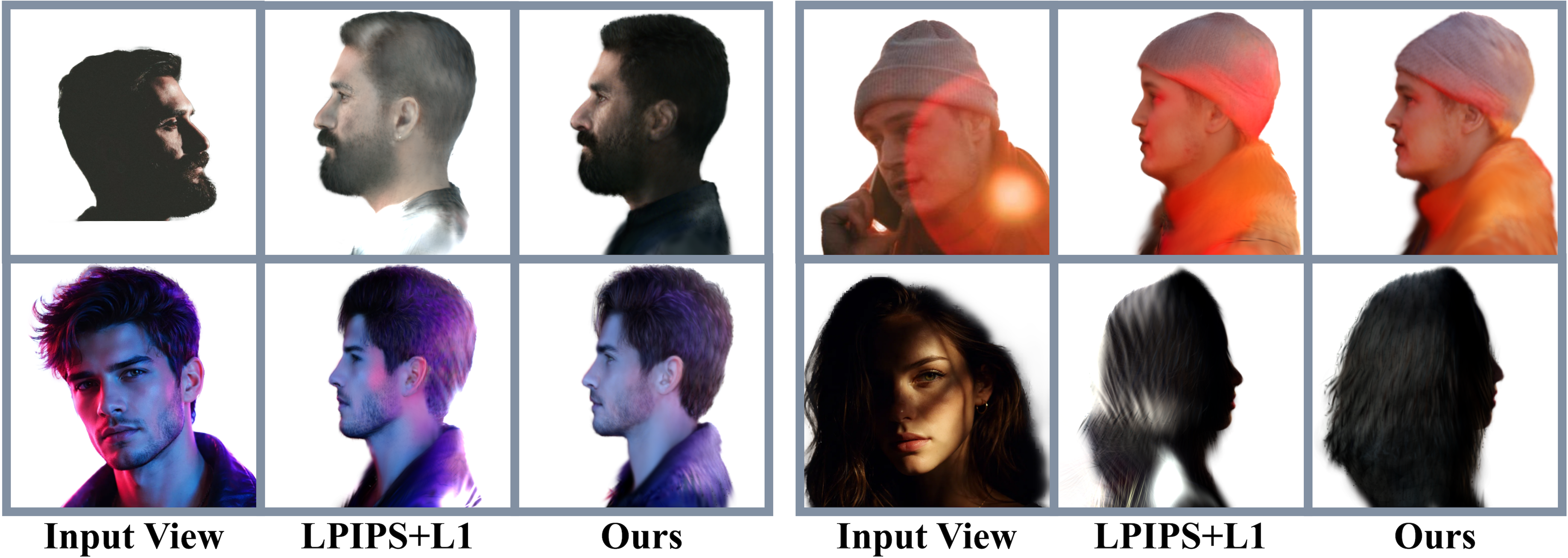}
  \caption{\textbf{Robustness of Our Model and an LPIPS+L1-Trained Ablation, Both Without the SHP-CNN.} Inputs are from UFDD \cite{ufdd} and SORA \cite{sora}.}
  \label{fig:x_rebuttal_lpips}
\end{figure*}

We compare our loss formulation and LPIPS+L1 in \cref{fig:x_rebuttal_lpips}, with deactivated SHP-CNN. Our loss shows strong robustness to difficult out-of-distribution inputs and viewing angles, while the standard loss leads to collapsed reconstructions.

\subsection{Computational Costs and Lightweight Model}
In \cref{tab:rebuttal_costs}, we compare the performance and computational requirements of our model and two smaller variants: (a) one model with four layers and no SHP-CNN, but using the same DINOv2 (Giant) encoder as our standard model, and (b) a model with the same configuration as (a), but using the smallest DINOv2 encoder. Both models provide strong performance and offer lightweight alternatives to our base model.

\begin{table}[h]
    \centering
    \scriptsize
    \setlength{\tabcolsep}{2.5pt}
    \renewcommand{\arraystretch}{1.15}

    \begin{tabular*}{\linewidth}{@{\extracolsep{\fill}}l||ccccc}
        \toprule
        \textbf{Ava-256 (Extreme Views)} & SSIM$\uparrow$ & LPIPS$\downarrow$ & DreamSim$\downarrow$ & ArcFace$\downarrow$ \\
        \midrule
        Light-S (4 layers, no SHP, DINOv2-S) & 15.4 & 0.313 & 0.120 & 0.316 \\
        Light (4 layers, no SHP, DINOv2-G) & \textbf{16.1} & 0.298 & 0.110 & 0.285 \\
        Ours (10 layers, SHP, DINOv2-G)     & 15.9 & \textbf{0.291} & \textbf{0.106} & \textbf{0.282} \\
        \bottomrule
    \end{tabular*}

    \begin{tabular*}{\linewidth}{@{\extracolsep{\fill}}l||cccc||ccc}
        \toprule
        \textbf{Model} & \textbf{DINOv2} & \textbf{Forward} & \textbf{GS} & \textbf{SHP} & \textbf{DINOv2} & \textbf{3D-Model} & \textbf{SHP} \\
        \midrule
        Light-S & 0.01\,s & 0.03\,s & 0.03\,s & - & 0.5\,GB & 2.1\,GB & - \\
        Light & 0.02\,s & 0.03\,s & 0.03\,s & - & 5.3\,GB & 2.1\,GB & - \\
        Ours & 0.02\,s & 0.07\,s & 0.03\,s & 0.04\,s & 5.3\,GB & 2.6\,GB & 6.1\,GB \\
        \bottomrule
    \end{tabular*}

   \caption{\textbf{Performance, computational cost, and memory requirements of our lightweight and standard model.}}
    \label{tab:rebuttal_costs}
\end{table}

\section{Disentangled 3D Editing}

\subsection{Additional Image- and Prompt-Based Editing Comparisons}
\begin{figure*}[t]
  \centering
  \includegraphics[width=\textwidth]{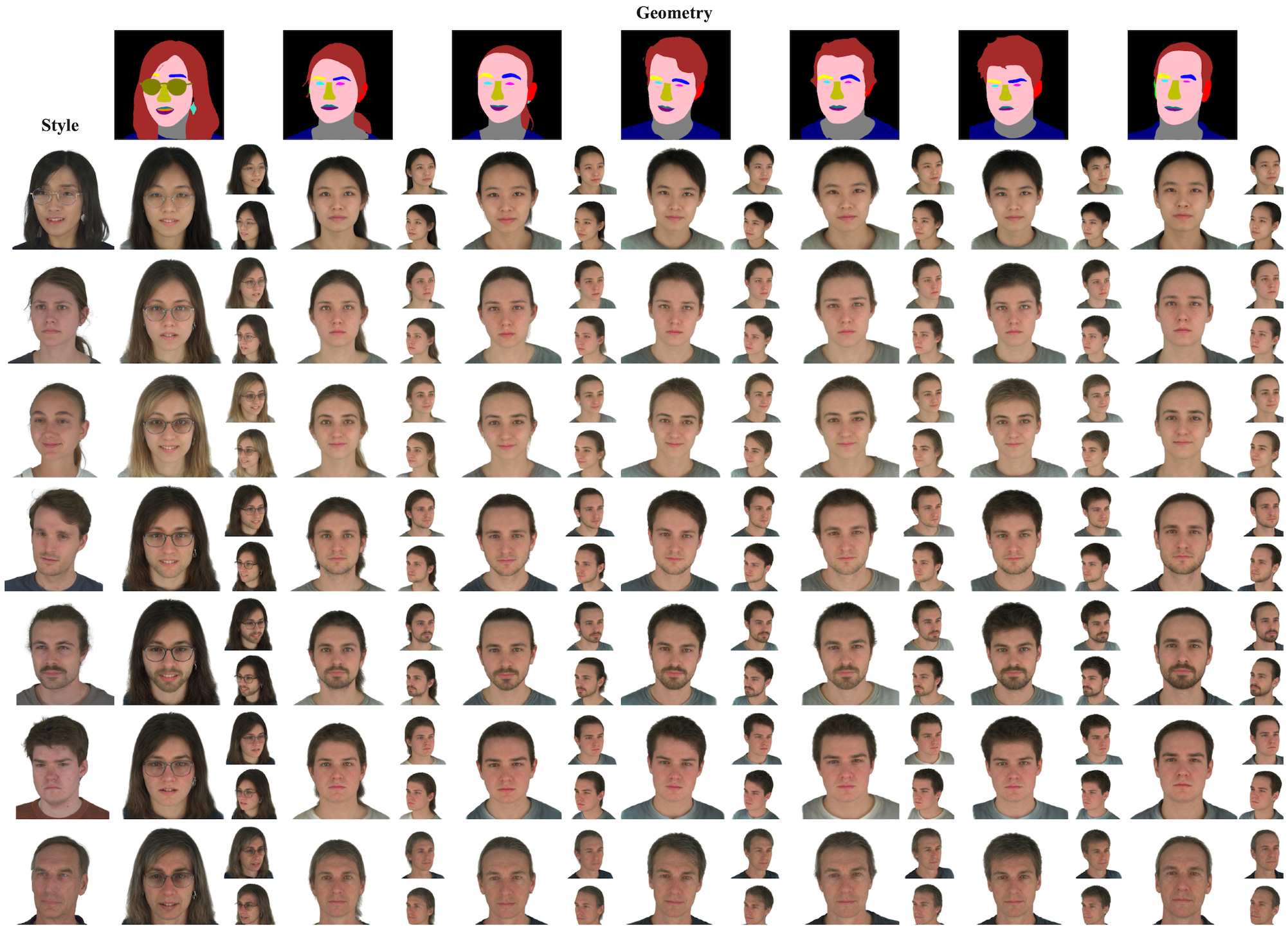}
  \caption{\textbf{Additional 3D Editing Results with Style Guided by Image.}}
  \label{fig:x_edit_img}
\end{figure*}

\begin{figure*}[t]
  \centering
  \includegraphics[width=0.85\textwidth]{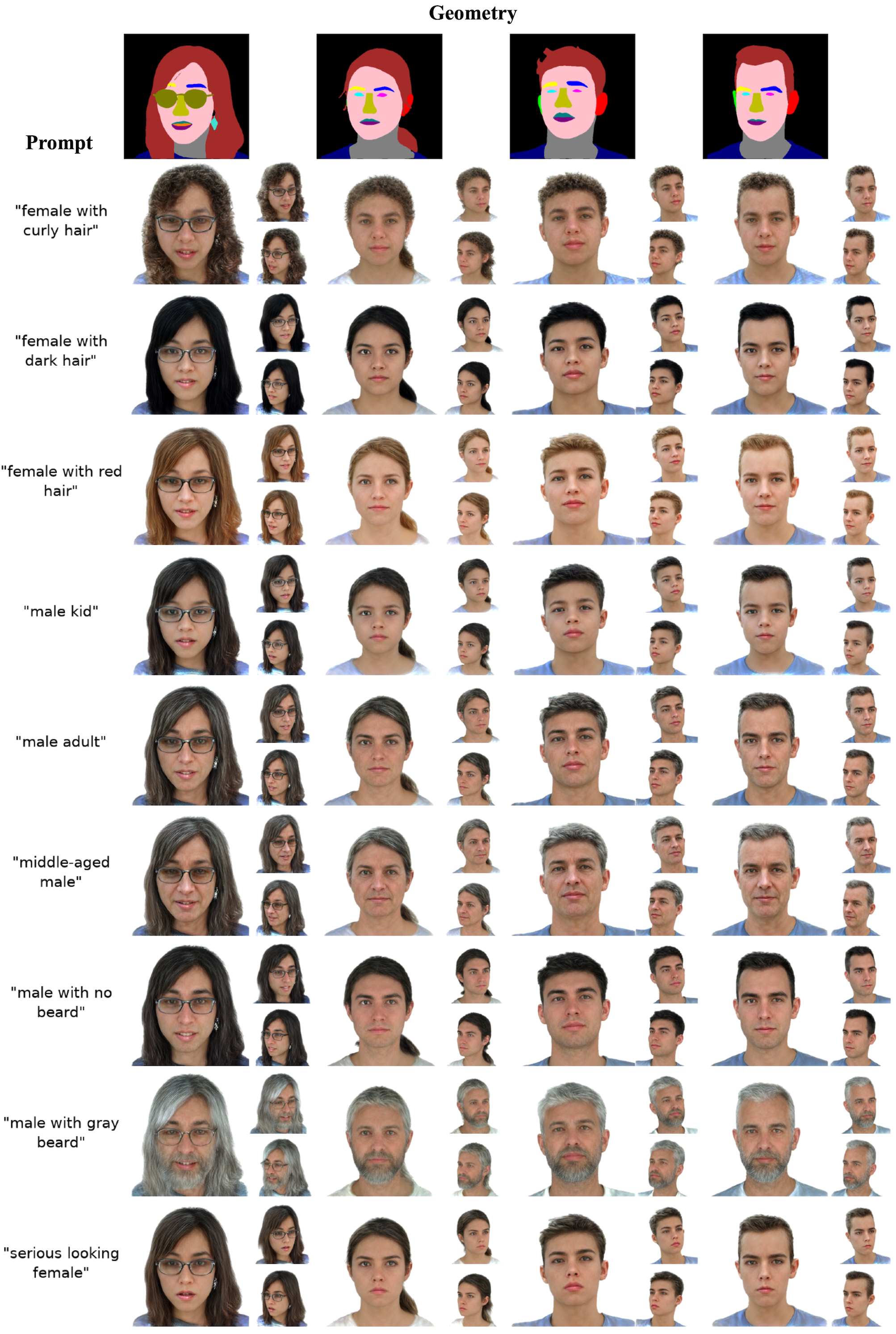}
  \caption{\textbf{Additional 3D Editing Results with Style Guided by Prompts.}}
  \label{fig:x_edit_prompt}
\end{figure*}

In \cref{fig:x_edit_img,fig:x_edit_prompt}, we present additional qualitative results for disentangled 3D editing using both image- and prompt-based stylization.

\subsection{3D Editing Web Application}
Our 3D editing web application exposes the full disentangled editing pipeline in an accessible interface. Users first upload an input image, from which the pipeline extracts a semantic segmentation map. The segmentation can then be interactively modified via drawing tools to, for example, reshape hair regions or add glasses.

Extracting a segmentation map from an image takes approximately 20 seconds, as it involves both GAGAvatar’s preprocessing~\cite{gagavatar} and FARL~\cite{farl} for semantic segmentation. This step is only required when uploading a new image; subsequent edits reuse the existing segmentation.

For stylization, users can either upload a reference image or provide a text prompt. Stylization with a reference image takes around 20 seconds, including CLIP-based feature extraction~\cite{clip} and a forward pass through our model. In contrast, generation using a text prompt is significantly faster, requiring only about 10 seconds. Overall, the web application demonstrates that our disentangled 3D editing framework is not only flexible but also practical for interactive use.

\section{Training \& Architecture Details}
\paragraph{Training.}
All models are trained on a single NVIDIA RTX~3090 GPU. The 3D reconstruction network is optimized for 15 epochs, corresponding to roughly 72~hours of training. Afterwards, the SHP-CNN head is trained for 30 epochs (about 10~hours) with the rest of the pipeline frozen, using only real data (i.e., excluding Cafca \cite{cafca}). The disentangled editing model reuses some of the pretrained 3D reconstruction weights and is further trained for 24 epochs (about 48~hours) without Cafca to avoid bias towards generating artificial heads. We employ the AdamW optimizer with a learning rate of $3\times 10^{-5}$ and a weight decay of $10^{-5}$.

\paragraph{Additional Architecture Details.}
We use a frozen DINOv2 ViT-g/14 encoder as a pre-processing stage, extracting four feature maps of dimensionality 1536 from layers 9, 19, 29, and 39 (similar to LAM \cite{lam}). The second encoder branch is a lightweight 2-layer ViT encoder with a dimensionality of 128. The decoder is a 10-layer ViT with a dimension of 1024 and operates on 4096 patches, each representing 16 Gaussians. We employ 16 cross-attention heads per decoder layer.

\section{Evaluation and Processing Details}
\subsection{Subjects used for quantitative evaluation:}
\noindent \textbf{NeRSemble \cite{nersemble}:} 059, 070, 370, 373, 374\\
\noindent \textbf{Ava-256 \cite{ava256}:}
\begin{itemize}
\item \texttt{20220809--1034--BJM420}
\item \texttt{20220815--1307--BMP511}
\item \texttt{20220831--0751--CMS162}
\item \texttt{20230224--1359--CMZ386}
\item \texttt{20230308--1352--BDF920}
\item \texttt{20230316--1103--BHK376}
\item \texttt{20230324--0820--AEY864}
\item \texttt{20230328--0800--BLY735}
\item \texttt{20230405--1635--AAN112}
\item \texttt{20230810--1630--ANX726}
\item \texttt{20230914--1105--BXQ083}
\end{itemize}

\paragraph{Cropping Alignment}  
\textbf{PanoHead} \cite{panohead} uses the tightest (smallest) image crops among all compared methods. To ensure fair and consistent evaluation across models, we applied the same PanoHead cropping to all methods for qualitative and quantitative comparison.

\paragraph{GAGAvatar Processing}
For \textbf{GAGAvatar} \cite{gagavatar}, their default rendering pipeline produces images with a black background. To standardize appearance and ensure comparability across methods, we replace the black background with white using their official \texttt{GAGAvatar Track} \cite{gagavatar} preprocessing pipeline.

\section{Note: Visualization of Intermediate Decoding States}
In our method overview (\cref{fig:method}), we visualize intermediate reconstruction states in our ViT decoder. For each visualization, we run a full forward pass, but control the activation of the cross-attention mechanisms. Specifically, to visualize the output after decoder layer $i$, we keep all cross-attention layers active up to and including layer $i$, while disabling cross-attention for all subsequent layers. Importantly, we retain the MLP blocks and skip connections in all layers, ensuring that feature propagation and refinement still occur. This setup allows us to isolate the contribution of 2D feature retrieval up to a specific depth in the decoder.

\end{document}